\documentclass[11pt]{article}

\PassOptionsToPackage{table}{xcolor}
\usepackage[preprint]{acl}

\usepackage{times}
\usepackage{latexsym}

\usepackage[T1]{fontenc}

\usepackage[utf8]{inputenc}

\usepackage{microtype}

\usepackage{inconsolata}

\usepackage{graphicx}

\usepackage{amsfonts}
\usepackage{nicefrac}
\usepackage{amsmath}

\usepackage[scale=1]{sourcecodepro}
\usepackage{float}
\usepackage{algorithm}
\usepackage{algpseudocode}
\usepackage{booktabs}
\usepackage{multirow}
\usepackage{tabularx}
\usepackage{array}
\usepackage{subcaption}

\usepackage{enumitem}
\usepackage{fontawesome5}
\usepackage{listings}
\usepackage[breakable,skins]{tcolorbox}
\setlist[itemize]{leftmargin=2em}
\setlist[enumerate]{leftmargin=2em}
\setlist[enumerate, 2]{leftmargin=1.5em}
\setlist[itemize, 2]{leftmargin=1.5em}

\title{\textsc{Frame}: Form-Restricted Authoring with Model-Based Evaluation}

\author{
  \textbf{Nan Jiang},
  \textbf{Mohammed Mohsin Hussain},
  \textbf{Onkar Tiwari},
  \textbf{Ankit Jain}\textsuperscript{\dag},
\\
  \textbf{Thong Q. Nguyen}\textsuperscript{\dag},
  \textbf{Abhishek Agarwal},
  \textbf{Vishal Chowdhary}
\\
  Microsoft Office AI
\\
  \small{
    \texttt{\{jiangnan,hussainmoh,Onkar.Tiwari,abhisag,vishalc\}@microsoft.com}
  }
\\
  \small{
    \texttt{kitsrishjain@gmail.com}, \texttt{thongnguyen1292@gmail.com}
  }
}

\definecolor{gold}{HTML}{FFD700}
\definecolor{silver}{HTML}{C0C0C0}
\definecolor{bronze}{HTML}{CD7F32}

\newcolumntype{L}{>{\raggedright\arraybackslash}X}
\newcolumntype{C}{>{\centering\arraybackslash}X}
\newcolumntype{R}{>{\raggedleft\arraybackslash}X}

\newcolumntype{f}{@{\extracolsep{\fill}}c}

\newcommand{\code}[1]{{\small\texttt{#1}}}

\renewcommand{\paragraph}[1]{\smallskip \noindent\textbf{#1}}

\newcounter{finding}
\renewcommand{\thefinding}{\arabic{finding}}
\newcommand{\finding}[1]{%
	\refstepcounter{finding}%
	\ifvmode
		\paragraph{Finding \thefinding: #1.}%
	\else
		\textbf{Finding \thefinding: #1.}\hspace{0.5em}%
	\fi
	\ignorespaces%
}

\definecolor{xmltag}{HTML}{1A6847}       
\definecolor{xmlval}{HTML}{A31515}       
\definecolor{codecomment}{HTML}{6A737D}  
\definecolor{jsonkey}{HTML}{0451A5}      
\definecolor{mdheading}{HTML}{0451A5}    
\definecolor{mdtablerule}{HTML}{6A737D}  
\definecolor{mdemph}{HTML}{7D4E24}       

\lstdefinestyle{exlst}{
  basicstyle=\ttfamily\fontsize{5.5pt}{6.5pt}\selectfont,
  breaklines=true,
  breakatwhitespace=false,
  breakindent=0pt,
  breakautoindent=false,
  columns=fullflexible,
  keepspaces=true,
  xleftmargin=0pt,
  frame=none,
  aboveskip=0pt,
  belowskip=0pt,
}

\lstdefinelanguage{iXBRL}{
  morestring=[b]",
  morecomment=[s]{<!--}{-->},
  morekeywords={ix:header,ix:resources,ix:nonFraction,
    xbrli:context,xbrli:period,xbrli:instant,
    xbrli:startDate,xbrli:endDate,xbrli:unit,xbrli:measure},
  alsoletter={:},
  sensitive=true,
}
\lstdefinestyle{xmlstyle}{
  style=exlst,
  language=iXBRL,
  keywordstyle=\color{xmltag},
  stringstyle=\color{xmlval},
  commentstyle=\color{codecomment}\itshape,
}

\lstdefinestyle{jsonstyle}{
  style=exlst,
  alsoletter={_},
  morekeywords={heading,children,type,objective,questions,columns,unit_id,unit_type,content,source_sentence,confidence,score,explanation,claim_id,claim,support_level},
  keywordstyle=\color{jsonkey},
  literate=
    *{"}{{{\color{black!55}{"}}}}1
     {\{}{{{\color{black!55}{\{}}}}{1}
     {\}}{{{\color{black!55}{\}}}}}{1}
     {[}{{{\color{black!55}{[}}}}{1}
     {]}{{{\color{black!55}{]}}}}{1}
     {...}{{{\color{codecomment}{.\kern0.5pt.\kern0.5pt.}}}}{3},
}

\lstdefinestyle{mdstyle}{
  style=exlst,
  moredelim=[il][\color{mdheading}]{\#\#\ },
  moredelim=[il][\color{mdheading}]{\#\ },
  literate=
    *{|}{{{\color{mdtablerule}{|}}}}{1}
     {---}{{{\color{mdtablerule}{-{}-{}-}}}}{3}
     {--}{{{\color{mdemph}{-{}-}}}}{2},
}

\lstdefinestyle{filestyle}{
  style=exlst,
  moredelim=[il][\color{mdheading}]{\#\#\ },
  moredelim=[il][\color{mdheading}]{\#\ },
  moredelim=[l][\color{codecomment}\itshape]{-----\ },
  literate=
    *{.md}{{{\color{jsonkey}{.md}}}}{3}
     {|}{{{\color{mdtablerule}{|}}}}{1},
}

\tcbset{
  exstyle/.style={
    fonttitle=\scriptsize\bfseries\rmfamily,
    coltitle=black,
    colbacktitle=gray!20,
    colback=gray!3,
    colframe=gray!45,
    boxrule=0.5pt,
    arc=2pt,
    top=3pt, bottom=2pt, left=4pt, right=4pt,
  }
}

\begin{document}

\maketitle

\begingroup
\renewcommand{\thefootnote}{\dag}
\footnotetext{Work done while at Microsoft.}
\endgroup

\begin{abstract}
Many professional reports reuse the structure of prior editions while updating the content. We formalize this setting as \emph{form-restricted report generation}: given a sample report and new grounding documents, generate a report that preserves the sample's form while accurately reflecting the new evidence. We introduce \textsc{Frame}, a benchmark of 120 EDGAR- and arXiv-derived instances across Financial Quarterly, Financial Annual, Academic Simple, and Academic Complex categories, built with a temporal three-snapshot pipeline. We evaluate outputs with three human-validated metrics: structural similarity via semantic tree-edit distance, coverage via information unit recall, and faithfulness via claim-level grounding verification. Across five LLMs, open-source models nearly match proprietary models on structure but lag on coverage and faithfulness; all models perform better on financial than academic reports, and performance declines with longer outputs and more complex grounding.
\end{abstract}

\section{Introduction}

Large language models (LLMs) can now produce fluent long-form documents \cite{openai2023gpt4,dubey2024llama3,zhao2023survey,koksal2023longform,que2024hellobenchevaluatinglongtext}, but many professional writing workflows are not open-ended. Analysts often reuse the form of an earlier report and update it with new evidence: quarterly earnings analyses, annual financial reviews, compliance reports, and recurring literature surveys all preserve structure while changing content. We call this setting \emph{form-restricted report generation}. It differs from standard summarization because the model must infer the target form from a sample report while grounding new content in separate evidence documents.

This setting is practically important and weakly covered by existing benchmarks. Financial filings such as 10-Qs and 10-Ks follow stable conventions while their data changes each period \cite{wu2023bloomberggpt,islam2023financebench}; survey-style academic reports similarly update an established organization as new papers appear \cite{wang2024autosurvey}. Existing long-form, summarization, grounded-generation, and data-to-document benchmarks \cite{koksal2023longform,que2024hellobenchevaluatinglongtext,fabbri2019multinews,lewis2021retrievalaugmentedgenerationknowledgeintensivenlp,wiseman2017challenges} do not directly test whether a model can transfer document form from one report while faithfully integrating new evidence. Surface metrics such as ROUGE and BERTScore also miss hierarchical organization and grounding errors \cite{zhang2019bertscore}.

We introduce \textsc{Frame} (\textbf{F}orm-\textbf{R}estricted \textbf{A}uthoring with \textbf{M}odel-Based \textbf{E}valuation), a benchmark of 120 instances across financial and academic domains. Each instance contains a sample report $S$, grounding documents $G$, and a gold report $Y$. A temporal construction pairs an earlier report with later evidence and target output, while a controlled synthesis pipeline keeps successive reports structurally aligned. We evaluate five LLMs with three human-validated metrics: structural similarity, information coverage, and grounding faithfulness.

Our contributions are as follows:
\begin{itemize}
    \item We formalize \emph{form-restricted report generation} as a distinct task and construct \textsc{FRAME}, a benchmark of 120 instances across financial and academic domains with controlled temporal structure.
    \item We develop a suite of evaluation metrics, i.e., structural similarity, coverage, and faithfulness, that go beyond surface-level comparison and aligns well with human preferences.
    \item We provide a comprehensive empirical analysis of five LLMs, revealing systematic performance patterns:
    \begin{enumerate}
        \item Open-source models nearly match proprietary models on structural similarity but lag substantially on coverage and faithfulness.
        \item All models perform better on financial reports than on academic reports.
        \item Performance degrades with increased task complexity, such as longer outputs, longer grounding contexts, and more complex structural requirements.
        \item Longer outputs mainly hurt structural quality and coverage, while faithfulness remains stable only for GPT-5.
        \item Longer grounding contexts reduce structural quality and coverage and increase faithfulness instability for weaker models.
        \item Reports with more tabular content tend to improve coverage and faithfulness, especially for open-source models.
    \end{enumerate}
\end{itemize}
\section{Related Work}

\subsection{LLM-Based Document Generation}
Document generation has progressed from templates and neural data-to-document systems \cite{wiseman2017challenges,shao-etal-2019-long} to retrieval-augmented and long-form LLM generation \cite{lewis2021retrievalaugmentedgenerationknowledgeintensivenlp}. Recent work adds structure through planning, schema post-processing, or style transfer \cite{shen-etal-2025-slot,ning2024skeletonofthoughtpromptingllmsefficient,wu2025implementinglongtextstyle}. These methods motivate structure-aware generation, but they do not benchmark whether a model can infer a target form from a prior report and populate it with new grounding evidence.

\subsection{Document Generation Evaluation}

Surface-overlap metrics such as BLEU and ROUGE often correlate poorly with human judgments, while embedding metrics improve semantic matching but miss discourse structure and factual consistency \cite{liu-etal-2023-g,zhang2019bertscore}. LLM-as-judge methods better capture holistic quality and factuality, but still require task-specific validation \cite{que2024hellobenchevaluatinglongtext,kulkarni-etal-2025-evaluating}.

Our evaluation combines three lines of work: structure metrics based on tree alignment \cite{DBLP:journals/corr/abs-1805-06869,liu2025generalizedtreeeditdistance}, faithfulness checks based on entailment, QA, and atomic facts \cite{durmus-etal-2020-feqa,Min2023FactScore}, and coverage metrics based on content units, nuggets, and semantic alignment \cite{nenkova2004pyramid,zhang2024qapyramid,zha2023alignscore}. \textsc{Frame} brings these dimensions together for form-restricted, evidence-grounded report generation.

\begin{figure*}[t]
    \centering
    \includegraphics[width=0.9\linewidth]{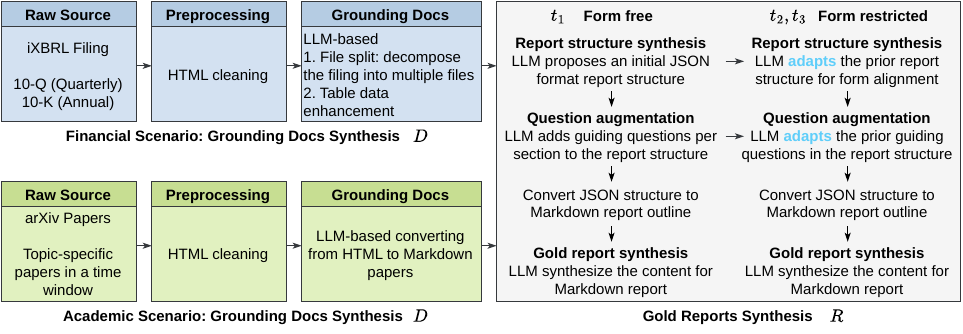}
    \caption{Overview of the \textsc{Frame} benchmark construction pipeline.}
    \label{fig:benchmark-pipeline}
\end{figure*}

\section{Benchmark Construction}
\label{sec:benchmark-construction}

We introduce \textsc{Frame}, a benchmark for \emph{form-restricted report generation} built from two complementary source families: financial filings from EDGAR and scientific papers from arXiv. This section formalizes the task and explains how benchmark instances are derived from temporally ordered snapshots (\S\ref{ssec:task-formulation}), describes the four categories and how they differ (\S\ref{ssec:source-selection}), details the domain-specific synthesis pipelines that produce grounding documents and gold reports (\S\ref{ssec:data-synthesis}), and summarizes the resulting statistics (\S\ref{ssec:benchmark-statistics}). Worked examples of the synthesis pipeline and a concrete benchmark instance are provided in Appendix~\ref{app:benchmark-examples}.

\subsection{Task Formulation}
\label{ssec:task-formulation}

Each benchmark instance is a triple $(S, G, Y)$: a \emph{sample report} $S$ that demonstrates the target form, \emph{grounding documents} $G$ with updated evidence, and a gold report $Y$. At inference time, the model receives $(S,G)$ and must generate $Y$, jointly transferring structure from $S$ and content from $G$. EDGAR filings and arXiv papers support this setup because both provide recurring reports whose form is relatively stable while content changes over time, and they offer complementary evidence modalities: financial data is table-dense, while academic data is figure- and prose-rich.

\paragraph{Benchmark instance construction.}
For each domain (a company or research topic), we collect three ordered snapshots $t_1<t_2<t_3$. At each timestamp, raw sources are converted into grounding documents $D_{t_i}$ and a gold report $R_{t_i}$. We then form two benchmark instances by shifting a window over adjacent snapshots:

{\small
\begin{align*}
(S_1, G_1, Y_1) &= (R_{t_1},\; D_{t_2},\; R_{t_2}), \\
(S_2, G_2, Y_2) &= (R_{t_2},\; D_{t_3},\; R_{t_3}).
\end{align*}
}

Thus $R_{t_i}$ provides the form exemplar, while $D_{t_{i+1}}$ and $R_{t_{i+1}}$ provide the updated evidence and target. This temporal pairing separates form reuse from content update and prevents $Y$ from being derived from $S$. Appendix~\ref{app:benchmark-examples} gives a concrete Adobe example.

\subsection{Source Selection and Category Design}
\label{ssec:source-selection}

Within the two source families we define four categories, each with 15 domains and 30 instances (yielding 120 instances in total; per-category statistics in Table~\ref{tab:benchmark-statistics}):

\begin{itemize}
\item \textbf{Financial Quarterly}: 10-Q filings covering a single quarter. The gold report updates recurring metrics (revenue, EPS, segment results) from a compact evidence set.

\item \textbf{Financial Annual}: 10-K filings covering a full fiscal year. Substantially longer and structurally richer than 10-Qs.

\item \textbf{Academic Simple}: Literature synthesis over a small set of arXiv papers on a topic (e.g., LoRA, LLM Agent); the gold report is a coherent survey of their key contributions.

\item \textbf{Academic Complex}: Literature synthesis over a larger paper set, requiring broader cross-document aggregation and richer multimodal understanding (more figures per grounding set).
\end{itemize}

\subsection{Data Synthesis Pipeline}
\label{ssec:data-synthesis}

Building valid benchmark instances requires (i)~constructing a diverse, multi-document grounding set $D_{t_i}$ at each timestamp, and (ii)~keeping the gold reports $R_{t_1}, R_{t_2}, R_{t_3}$ within a domain form-aligned so an earlier report is informative as a sample. Figure~\ref{fig:benchmark-pipeline} illustrates the full pipeline; we describe domain-specific grounding construction first, then the shared report synthesis.

\paragraph{Financial grounding documents.}
Each iXBRL filing is converted to HTML with XBRL namespaces, inline styles, and scripts stripped. An LLM then (1)~\emph{file-splits} the cleaned HTML into semantically coherent units (business overviews, risk factors, segment results, financial statements, etc.), and (2)~\emph{enhances tabular content} into well-formatted Markdown so numerical evidence is accessible. The resulting set mixes prose- and table-centric documents. \textbf{Financial Annual} applies the same pipeline to longer, richer 10-K filings.

\paragraph{Academic grounding documents.}
Each arXiv paper in a time window is HTML-cleaned and converted by an LLM into structured Markdown, preserving equations and tables. \textbf{Academic Complex} uses more papers per window than \textbf{Simple}.

\paragraph{Gold report synthesis.}
We synthesize $R_{t_i}$ in four steps, applied identically to both source families: (1)~\emph{Report structure synthesis}, producing a hierarchical JSON of sections and subsections with element-level objectives; (2)~\emph{Question augmentation}, adding 5--8 guiding questions per section; (3)~\emph{Convert to Markdown outline}, rendering the JSON into a template with per-section placeholders; (4)~\emph{Gold report synthesis}, filling each placeholder by prompting an LLM with the full grounding set, the complete structure, and the per-section questions.

Steps (1) and (2) differ only in initialization. For $t_1$ (\emph{form free}) the LLM proposes a structure from scratch based on $D_{t_1}$; for $t_2$ and $t_3$ (\emph{form restricted}) it adapts the prior report's structure to the new grounding. This template-conditioning is what keeps successive snapshots form-aligned and makes the sample report informative. Appendix~\ref{app:benchmark-examples} Figure~\ref{fig:synthesis-example} walks through the pipeline on Adobe's Q1~FY2025 10-Q. The full text of every prompt used by the grounding-document and gold-report stages is reproduced in Appendix~\ref{app:benchmark-prompts}.

\begin{table*}[t]
\centering
\scriptsize
\setlength{\tabcolsep}{3pt}
\caption{Benchmark statistics per category. All counts are instance-level averages over the 30 instances in each category. Token counts use the \code{o200k\_base} tokenizer.}
\label{tab:benchmark-statistics}
    \begin{tabular*}{\linewidth}{l@{\extracolsep{\fill}}ccccccccccc}
    \toprule
    & & \multicolumn{4}{c}{\textbf{Grounding Documents}} & \multicolumn{3}{c}{\textbf{Sample Report}} & \multicolumn{3}{c}{\textbf{Gold Report}} \\
    \cmidrule(lr){3-6} \cmidrule(lr){7-9} \cmidrule(lr){10-12}
    \textbf{Category} & \textbf{Instances} & \textbf{Files} & \textbf{Tokens} & \textbf{Tables} & \textbf{Images} & \textbf{Tokens} & \textbf{Sec.} & \textbf{Tables} & \textbf{Tokens} & \textbf{Sec.} & \textbf{Tables} \\
    \midrule
    Financial Qtr.   & 30 & 11.2 & 24.2k & 39.6 & ---  & 9.1k  & 12.5 & 12.2 & 8.8k  & 12.5 & 12.2 \\
    Financial Ann.   & 30 & 20.8 & 59.2k & 70.8 & ---  & 22.1k & 18.9 & 33.7 & 22.4k & 19.1 & 32.6 \\
    Academic Simp.   & 30 & 3.4  & 28.0k & 12.9 & 13.7 & 21.6k & 19.1 & 8.2  & 22.6k & 18.8 & 8.6  \\
    Academic Comp.   & 30 & 6.6  & 57.1k & 28.8 & 30.4 & 19.3k & 21.8 & 7.1  & 20.8k & 22.0 & 7.4  \\
    \bottomrule
    \end{tabular*}
\end{table*}

\subsection{Benchmark Statistics}
\label{ssec:benchmark-statistics}

Table~\ref{tab:benchmark-statistics} reports per-category averages across the 120 instances.

\paragraph{Input scale.} Grounding token counts range from 24k (Financial Quarterly) to 59k (Financial Annual and Academic Complex), spanning more than a $2\times$ difference in input size. Financial grounding sets are table-dense (40--71 tables on average), while academic grounding sets are figure-rich (14--30 images), reflecting the contrasting evidence modalities of the two source families.

\paragraph{Output scale and structure.} Gold reports range from 9k tokens (Financial Quarterly) to 23k tokens (Financial Annual). Structural complexity varies from 12.5 sections and 12.2 tables per report (Financial Quarterly) to 22.0 sections and 7.4 tables (Academic Complex), contrasting the table-centric format of financial reports with the prose-heavy structure of academic syntheses.

\paragraph{Sample--gold alignment.} Within each category, sample and gold reports have closely matching token counts, section counts, and table counts (Table~\ref{tab:benchmark-statistics}). This confirms that the template-conditioned synthesis pipeline successfully preserves report form across successive snapshots, making the earlier report a meaningful structural exemplar.

\section{Evaluation Metrics}
\label{sec:metrics}

We evaluate generated reports along three complementary dimensions: structural similarity to the gold report (\S\ref{sec:structural-similarity}), coverage of the gold report's information (\S\ref{sec:coverage}), and faithfulness to the grounding documents (\S\ref{sec:faithfulness}). We then validate all three metrics against human judgments (\S\ref{sec:human-alignment}).

\subsection{Structural Similarity}
\label{sec:structural-similarity}
We extract each report into an ordered structure tree whose nodes represent sections, subsections, tables, figures, or bullet groups. We then align the generated and gold trees with a semantic tree-edit distance: node substitutions use embedding-based semantic similarity, while insertion and deletion handle missing or extra structure. The final score is a normalized similarity in $[0,1]$.

\paragraph{Node similarity.}
Each node stores its type, title, objective description, and ordered children. An LLM extracts this tree in JSON format. For two nodes $r$ and $s$, semantic similarity combines node type and objective embeddings:

{\small
\[
\begin{aligned}
\mathrm{sim}(r,s)={}&w\cdot \mathrm{proj}\!\left(\cos(e^r_{\mathrm{type}},e^s_{\mathrm{type}})\right) \\
&+(1-w)\cdot \mathrm{proj}\!\left(\cos(e^r_{\mathrm{obj}},e^s_{\mathrm{obj}})\right).
\end{aligned}
\]
}

The projection function maps raw cosine similarity to $[0,1]$; we compare linear, power, and RBF families during human-alignment tuning (\S\ref{sec:human-alignment-structural}) and use RBF with $\delta=0.10$ and $w=0.2$ in the final experiments. Node substitution cost is $2(1-\mathrm{sim}(r,s))$, so a fully dissimilar substitution matches one deletion plus one insertion. The normalized tree-alignment score is reported as structural similarity.

\paragraph{Notations.}
Let $T$ denote an ordered rooted tree and let $\mathrm{Root}(T)$ be its root node. For a node $r$, let $\mathrm{Ch}(r)$ be its ordered list of children. We write $\mathrm{SubT}(r)$ for the ordered forest formed by the subtrees rooted at each child of $r$, that is,
$\mathrm{SubT}(r)=\{T_c \mid c \in \mathrm{Ch}(r)\}$, in the same order as $\mathrm{Ch}(r)$.
A \emph{forest} $F$ is an ordered list of trees, written $F=\{T_1,\ldots,T_m\}$; we use $\mathrm{First}(F)$ and $\mathrm{Rest}(F)$ to denote its first tree and the remaining list. Finally, $\mathrm{Size}(T)$ is the number of nodes in tree $T$. For depth-based variants, $\mathrm{depth}(r)$ denotes the depth of node $r$ in its tree (with the root at depth $0$).

\paragraph{Similarity algorithm.}
We compute a semantic tree-edit-style distance by aligning two trees via (i) a \emph{node substitution cost} and (ii) a \emph{forest alignment cost} over their child subtrees, then normalize the distance into a similarity score.

\begin{algorithm}[t]
\footnotesize
\caption{Structural similarity via semantic tree/forest alignment}
\label{alg:struct-sim}
\begin{algorithmic}[1]
    \Function{CostN}{$r, s$} \Comment{node substitution cost}
        \State \Return $2 \cdot \left(1 - \mathrm{semantic\_similarity}(r,s)\right)$
    \EndFunction
    \Statex

    \Function{CostT}{$T_r, T_s$} \Comment{tree alignment cost}
        \State $r \gets \mathrm{Root}(T_r)$; \;\; $s \gets \mathrm{Root}(T_s)$
        \State \Return $\Call{CostN}{r,s} + \Call{CostF}{\mathrm{SubT}(r), \mathrm{SubT}(s)}$
    \EndFunction
    \Statex

    \Function{CostF}{$F_1, F_2$} \Comment{forest alignment cost}
        \If{$F_1 = \emptyset$}
            \State \Return $\sum_{T \in F_2} \mathrm{Size}(T)$ \Comment{insert all remaining nodes}
        \EndIf
        \If{$F_2 = \emptyset$}
            \State \Return $\sum_{T \in F_1} \mathrm{Size}(T)$ \Comment{delete all remaining nodes}
        \EndIf

        \State $T \gets \mathrm{First}(F_1)$; \;\; $U \gets \mathrm{First}(F_2)$
        \State $F_1' \gets \mathrm{Rest}(F_1)$; \;\; $F_2' \gets \mathrm{Rest}(F_2)$
        \State $r \gets \mathrm{Root}(T)$; \;\; $s \gets \mathrm{Root}(U)$

        \State $c_{\text{sub}} \gets \Call{CostT}{T, U} + \Call{CostF}{F_1', F_2'}$
        \State $c_{\text{del}} \gets 1 + \Call{CostF}{\mathrm{SubT}(r) \cup F_1', F_2}$
        \State $c_{\text{ins}} \gets 1 + \Call{CostF}{F_1, \mathrm{SubT}(s) \cup F_2'}$

        \State \Return $\min\left(c_{\text{sub}}, c_{\text{del}}, c_{\text{ins}}\right)$
    \EndFunction
    \Statex

    \Function{StructuralSimilarity}{$T_1, T_2$}
        \State $d \gets \Call{CostT}{T_1, T_2}$
        \State \Return $1 - \dfrac{d}{\mathrm{Size}(T_1) + \mathrm{Size}(T_2)}$
    \EndFunction
\end{algorithmic}
\end{algorithm}

\paragraph{Design rationale.}
The algorithm is based on classic ordered tree edit distance, which aligns two trees by first aligning their roots and then aligning their child \emph{forests}. The three cost functions in Algorithm~\ref{alg:struct-sim} are central to this process. $\mathrm{CostN}(r,s)$ is the cost of treating node $r$ (from the first tree) and node $s$ (from the second tree) as corresponding. $\mathrm{CostT}(T_r,T_s)$ is the cost of aligning the entire trees rooted at $r$ and $s$, defined as the root node cost plus the cost of aligning their child forests. $\mathrm{CostF}(F_1,F_2)$ aligns two ordered lists of subtrees and returns the minimum cost to make the forests match.

Forest alignment supports three operations. \emph{Substitution} aligns the first tree in each forest and recurses on the remaining suffixes. \emph{Deletion} removes the root of the first tree in $F_1$ at unit cost and promotes its children to the forest frontier (equivalently, deleting a node while retaining its descendants). \emph{Insertion} is symmetric: it removes the root of the first tree in $F_2$ at unit cost and promotes its children. Taking the minimum over these three choices yields a dynamic program analogous to sequence edit distance, lifted to forests of subtrees.


\begin{figure*}[!t]
\centering

\begin{minipage}[t]{0.495\linewidth}

\begin{tcolorbox}[exstyle, equal height group=struct, title={Gold Report Structure $T_{\text{gold}}$}]
\begin{lstlisting}[style=jsonstyle]
{
  "heading": "Company Consolidated Analytical Report (Q[XX] FY[XX])",
  "type": "section",
  "objective": "Provide a comprehensive, templatized structure for a company's quarterly analytical report summarizing performance, financials...",
  "children": [
    {
      "heading": "1. Executive Summary and Key Highlights",
      "type": "section",
      "objective": "Summarize the period's key results, performance drivers, notable events, foreign exchange impacts, and liquidity position...",
      "children": [ ... ]
    },
    {
      "heading": "2. Business Model and Segment Performance",
      "type": "section",
      "objective": "Describe business model, key offerings and platforms, customer segments, go-to-market themes, and summarize segment and geographic performance.",
      "children": [ ... ]
    },
    ...
    {
      "heading": "4. Financial Statements Overview",
      "type": "section",
      "objective": "Provide summarized income statement, balance sheet, and cash flow information with analytical commentary on drivers and comparability.",
      "children": [
        {
          "heading": "4.1 Income Statement",
          "children": [
            { "type": "text", ... },
            { "type": "table", ... },
            { "type": "text", ... }
          ]
        },
        { "heading": "4.2 Balance Sheet", ... },
        { "heading": "4.3 Cash Flow", ... }
      ]
    },
    { "heading": "5. Liquidity ...", ... },
    ...
}
\end{lstlisting}
\end{tcolorbox}

\end{minipage}%
\hfill%
\begin{minipage}[t]{0.495\linewidth}

\begin{tcolorbox}[exstyle, equal height group=struct, title={GPT-5 Generated Report Structure $T_{\text{gen}}$}]
\begin{lstlisting}[style=jsonstyle]
{
  "heading": "Consolidated Analytical Report for [Company] Q[X] FY[XX]",
  "type": "section",
  "objective": "A templatized structure to present a company's quarterly performance, financials, liquidity, capital structure, risks, and outlook...",
  "children": [
    {
      "heading": "1. Executive Summary and Key Highlights",
      "type": "section",
      "objective": "Provide a concise overview of the quarter, including headline revenue and mix, segment and geographic highlights, profitability and EPS...",
      "children": [ ... ]
    },
    {
      "heading": "2. Business Context and Segment Performance",
      "type": "section",
      "objective": "Describe the operating model, portfolio of products and services, innovation focus (e.g., AI capabilities), and performance by business segment and geography.",
      "children": [ ... ]
    },
    ...
    {
      "heading": "4. Financial Statements Overview",
      "type": "section",
      "objective": "Summarize key results from the income statement, balance sheet, and cash flows, including drivers of year-over-year changes.",
      "children": [
        { "type": "text", ... },
        { "type": "table", ... },
        { "type": "text", ... },
        { "type": "text", ... },
        { "type": "table", ... },
        { "type": "text", ... },
        ...
      ]
    },
    { "heading": "5. Liquidity ...", ... },
    ...
}
\end{lstlisting}
\vspace{21pt}
\end{tcolorbox}

\end{minipage}

\caption{%
  Extracted document structures for the Adobe Q2~FY2025 gold report ($T_{\text{gold}}$, left) and the GPT-5 generated report ($T_{\text{gen}}$, right). The \textsc{StructuralSimilarity} algorithm aligns these two trees to produce a similarity score.%
}
\label{fig:structure-example}
\end{figure*}

Figure~\ref{fig:structure-example} illustrates the two structure trees that the algorithm takes as input, using the Adobe Q2~FY2025 instance as a running example. The gold report structure $T_{\text{gold}}$ (left) and the GPT-5 generated report structure $T_{\text{gen}}$ (right) share all eleven top-level sections and largely agree on child node types and ordering. However, they exhibit two representative kinds of divergence that the algorithm must handle. First, headings differ in surface form while conveying the same meaning (e.g., ``Business Model and Segment Performance'' vs.\ ``Business Context and Segment Performance''), requiring semantic rather than exact matching. Second, the gold structure organizes Section~4 into three nested subsections (4.1--4.3), whereas GPT-5 places the same content as a flat sequence of text and table nodes, requiring the deletion operation to promote children when aligning the two forests.

\paragraph{Depth-weighted variant (optional).}
Top-level structural differences, such as missing or reordered major sections, are typically more salient to human readers than disagreements in deeply nested details. To capture this intuition, we optionally replace the flat substitution cost $\mathrm{CostN}(r, s)$ with a depth-weighted variant:

{\small
\[
\mathrm{CostN}_{\text{depth}}(r,s)
\;=\; \frac{2\,\bigl(1-\mathrm{sim}(r,s)\bigr)}{(d_{\text{eff}}+1)^{\gamma}}
\]
}

where $\gamma > 0$ is a hyperparameter and

{\small
\[
d_{\text{eff}} \;=\; \min\bigl(\mathrm{depth}(r),\,\mathrm{depth}(s)\bigr).
\]
}

The denominator $(d_{\text{eff}}+1)^{\gamma}$ decreases the substitution cost as nodes become deeper: a mismatch at the root ($d_{\text{eff}}=0$) retains the full cost of $2\,(1-\mathrm{sim}(r,s))$, while mismatches at greater depths are progressively discounted. We take the minimum depth of the two nodes so that if either node is shallow, the substitution is still penalized strongly. Larger $\gamma$ amplifies this depth sensitivity, while $\gamma=0$ recovers the unweighted formulation.

This is a drop-in modification: the tree and forest alignment recurrences remain unchanged, and only the per-node substitution term is rescaled, allowing the same dynamic program to be reused.

\subsection{Coverage}
\label{sec:coverage}
Coverage measures whether the generated report preserves salient information from the gold report. An LLM extractor decomposes the gold report into atomic information units $\mathcal{U}_{\mathrm{ref}}=\{u_1,\ldots,u_N\}$, covering key facts, figures, and findings section by section. We sample 100 units for tractability. For each sampled unit, an LLM judge searches the generated report for semantically matching content and assigns a graded score $c_i\in[0,1]$, where 0 means absent, 1 means fully covered, and intermediate values denote partial coverage. We define Information Unit Recall as

{\small
\[
\mathrm{IUR}=\frac{1}{|\mathcal{U}_{\mathrm{ref}}|}\sum_{u_i\in\mathcal{U}_{\mathrm{ref}}}c_i.
\]
}

Worked examples of information-unit extraction and coverage judgments are given in Appendix~\ref{app:coverage-faithfulness-examples}.

\subsection{Faithfulness}
\label{sec:faithfulness}
Faithfulness measures whether claims in the generated report are supported by the grounding documents. An LLM extractor decomposes the generated report into verifiable claims $\mathcal{C}=\{x_1,\ldots,x_M\}$, each expressing a single checkable proposition. We sample 100 claims and assign each a support score $s_j\in[0,1]$ against the grounding documents. The verifier first checks direct support from explicit evidence; claims scoring below 0.75 are passed to a deeper verification stage that considers logical entailment under reasonable assumptions. Report-level faithfulness is

{\small
\[
\mathrm{Faithfulness}=\frac{1}{|\mathcal{C}|}\sum_{x_j\in\mathcal{C}}s_j.
\]
}

Worked examples of claim extraction and verification are given in Appendix~\ref{app:coverage-faithfulness-examples}.

\subsection{Alignment with Human Judgment}
\label{sec:human-alignment}

To validate our automated evaluation approaches, we randomly sampled 10 data points: 5 from the Financial Ann.\ category and 5 from the Financial Qtr.\ category. For each sampled data point, we prompted the five studied models to generate a report, and then performed human evaluation and alignment analysis on the resulting outputs.

\subsubsection{Structural Similarity}
\label{sec:human-alignment-structural}
For structural similarity, we evaluate our algorithm by measuring how well its induced ranking correlates with human preference rankings. In this pipeline, report structures are extracted using GPT-5 with high reasoning effort, and node embeddings for semantic similarity are obtained using \code{text-embedding-small}. Rather than asking annotators to assign absolute similarity scores, which is cognitively demanding and prone to inconsistency, we ask them to rank the five model-generated reports for each data point from best to worst based on structural quality. We then compute the same ranking using our \textsc{StructuralSimilarity} algorithm and measure agreement between the two rankings using Spearman $r$ and Kendall $\tau$.

\begin{figure*}[tp]
    \centering
    \begin{minipage}[c]{0.5\linewidth}
        \centering
        \includegraphics[width=\linewidth]{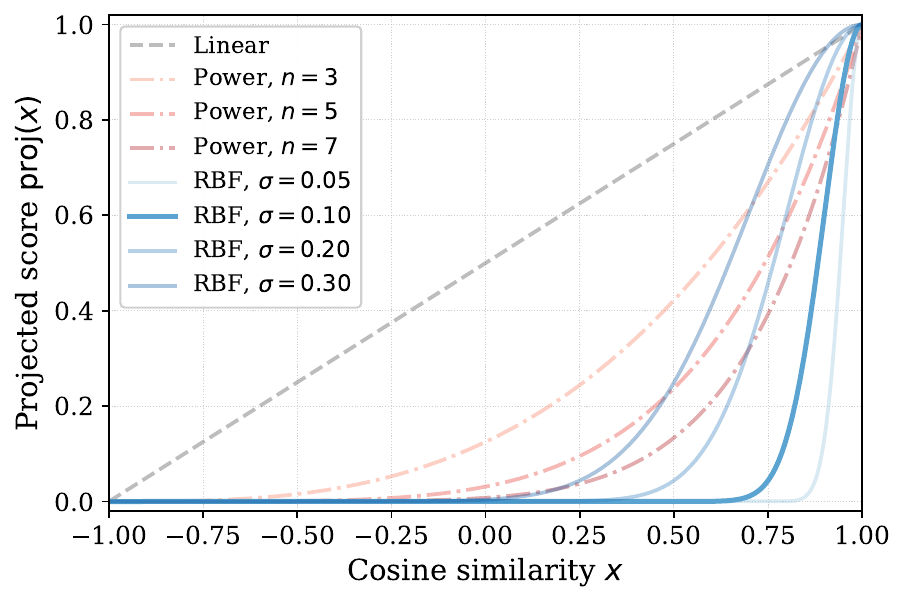}
        \captionof{figure}{Studied projection functions mapping cosine similarity to calibrated semantic scores.}
        \label{fig:score_projection_function}
    \end{minipage}\hfill
    \begin{minipage}[c]{0.46\linewidth}
        \centering
        \scriptsize
        \captionof{table}{Ranking correlation between the \textsc{StructuralSimilarity} algorithm and human preference across different projection functions. Type weight $w=0.2$, no depth adjustment.}
        \label{tab:aggregation_correlation}
        \begin{tabular*}{\linewidth}{@{\extracolsep{\fill}}llcc}
        \toprule
        \textbf{Type} & \textbf{Param} & \textbf{Spearman $r$} & \textbf{Kendall $\tau$} \\
        \midrule
        Linear & --          & 0.590 & 0.491 \\
        \midrule
        \multirow{3}{*}{Power} & $n=3$ & 0.710 & 0.594 \\
                               & $n=5$ & 0.720 & 0.608 \\
                               & $n=7$ & 0.720 & 0.608 \\
        \midrule
        \multirow{4}{*}{RBF}   & $\delta=0.05$ & 0.740 & 0.632 \\
                               & $\delta=0.10$ & \textbf{0.780} & \textbf{0.661} \\
                               & $\delta=0.20$ & 0.750 & 0.624 \\
                               & $\delta=0.30$ & 0.710 & 0.590 \\
        \bottomrule
        \end{tabular*}
    \end{minipage}
\end{figure*}

\paragraph{Projection Function.}
Table~\ref{tab:aggregation_correlation} reports ranking correlations across different projection functions, using a fixed type weight $w=0.2$ and no depth adjustment. The results show that the choice of projection function meaningfully affects alignment with human preference. The linear projection yields the weakest correlation, while both power and RBF projections improve over this baseline. The RBF kernel with $\delta=0.10$ achieves the best correlation with human rankings ($r=0.780$, $\tau=0.661$). We attribute this to the tendency of cosine similarities between text embeddings to concentrate in a high range, leaving little variation to distinguish between nodes that are genuinely similar and those that merely share surface-level vocabulary. Figure~\ref{fig:score_projection_function} visualizes these projection functions and shows that the RBF mapping places stronger emphasis on high-similarity regions near $x=1$ ($x$ being the cosine similarity). The RBF kernel's sharply peaked response near $x=1$ amplifies these fine-grained differences, producing a more discriminative score that better reflects human perceptions of structural similarity. We therefore adopt RBF with $\delta=0.10$ as the default projection function in all subsequent experiments.

\begin{table*}[tp]
\scriptsize
\begin{minipage}[t]{0.52\linewidth}
    \centering
    \captionof{table}{Ranking correlation between \textsc{StructuralSimilarity} and human preference under different type weights $w$ (RBF projection fixed, $\delta=0.10$).}
    \label{tab:type_weight_correlation}
    \begin{tabular*}{\linewidth}{@{\extracolsep{\fill}}ccc}
    \toprule
    \textbf{$w$} & \textbf{Spearman $r$} & \textbf{Kendall $\tau$} \\
    \midrule
    0.1 & 0.760 & 0.631 \\
    0.2 & \textbf{0.780} & \textbf{0.661} \\
    0.3 & 0.770 & 0.647 \\
    0.5 & 0.730 & 0.632 \\
    0.7 & 0.720 & 0.613 \\
    \bottomrule
    \end{tabular*}
\end{minipage}\hfill
\begin{minipage}[t]{0.44\linewidth}
    \centering
    \captionof{table}{Ranking correlation under different depth-adjustment strengths $\gamma$ (RBF projection, $\delta=0.10$, $w=0.2$).}
    \label{tab:depth_adjustment_correlation}
    \begin{tabular*}{\linewidth}{@{\extracolsep{\fill}}lccc}
    \toprule
    \textbf{Setting} & \textbf{$\gamma$} & \textbf{$r$} & \textbf{$\tau$} \\
    \midrule
    No depth adjustment & -- & 0.780 & 0.661 \\
    \hline
    \multirow{4}{*}{Depth adjustment} & 0.3 & 0.770 & 0.647 \\
     & 0.5 & 0.770 & 0.647 \\
     & 1.0 & 0.780 & \textbf{0.666} \\
     & 2.0 & \textbf{0.790} & 0.658 \\
    \bottomrule
    \end{tabular*}
\end{minipage}
\end{table*}

\paragraph{Type Weight.}
After fixing the projection function to RBF with $\delta=0.10$, we analyzed the effect of the type-weight hyperparameter $w$ in
\textsc{StructuralSimilarity}. We tested {\small $w\in\{0.1, 0.2, 0.3, 0.5, 0.7\}$} and measured rank correlation with human preference. As shown in Table~\ref{tab:type_weight_correlation}, performance peaks at {\small $w=0.2$} (Spearman {\small $r=0.780$}, Kendall {\small $\tau=0.661$}), while both smaller and larger values lead to lower alignment.

This trend suggests that assigning a moderate weight to node type and a larger weight to node objective text yields the best agreement with human judgment. When $w$ becomes large (e.g., $0.5$ or $0.7$), the model overemphasizes coarse structural type matching at the expense of semantic content, reducing discriminative power.

\paragraph{Depth Adjustment.}
We further evaluated the optional depth-adjusted substitution cost under the best non-depth setting found above (RBF projection with {\small $\delta=0.10$} and type weight {\small $w=0.2$}). We compare a no-depth-adjustment baseline against depth-adjusted variants with {\small $\gamma\in\{0.1,0.3,0.5,1.0,2.0\}$}. Table~\ref{tab:depth_adjustment_correlation} shows that depth weighting has only a limited effect on alignment with human preference: Spearman $r$ varies within a narrow range (0.770--0.790), and Kendall $\tau$ remains similarly stable (0.647--0.666).

Although a few depth-weighted variants slightly improve one correlation metric, the gains are small and not consistently reflected in both metrics. We therefore do not use depth adjustment in the default configuration and retain the simpler unweighted formulation.

\subsubsection{Coverage}
\label{sec:human-alignment-coverage}
For each of the 10 sampled data points, we randomly selected 50 information units extracted by the LLM judge from the ground-truth reports, resulting in 500 information units in total, each with an LLM-judge score indicating how well the information unit is covered in the model-generated report. For this metric, we use GPT-5 with high reasoning effort for both information-unit extraction and coverage validation.

Each information unit was then evaluated by two human annotators, who assigned coverage scores on a three-point scale: 0 for not covered in the model-generated report, 50 for partially covered, and 100 for covered. We use this three-point scale for human annotators, rather than the same continuous 0--100 scale used by the LLM judge, because humans are generally more reliable and consistent when making categorical judgments than when assigning fine-grained numerical scores.

These human judgments were then compared with the continuous scores produced by our automated LLM judge. Because the comparison involves continuous LLM-generated scores and discrete ordinal human ratings, we use multiple complementary metrics to assess inter-rater agreement. We report:
\begin{itemize}
    \item \textbf{Percentage Agreement} ($\mathrm{PA}$)~\cite{pa}, which measures the proportion of judgments where the absolute difference between human and LLM scores falls within a specified threshold. Scores range from 0--100\%, with higher values indicating more frequent direct alignment. We report $\mathrm{PA}@0$ (exact agreement) and $\mathrm{PA}@25$ (agreement within one rating level) because $\mathrm{PA}@25$ matches the granularity of averaged human ratings (0, 25, 50, 75, 100 after averaging two annotators' scores), making it particularly meaningful for practical alignment in our setting.

    \item \textbf{Quadratic Weighted Kappa} ($\mathrm{QWK}$)~\cite{qwk}, a metric designed for ordinal categorical data that accounts for the magnitude of disagreement by assigning quadratically increasing penalties to larger discrepancies while correcting for chance agreement. Reported on a 0--1 scale, higher scores indicate stronger agreement beyond chance. $\mathrm{QWK}$ is particularly appropriate for our setting, where disagreement between scores of 0 and 25 should be penalized less than disagreement between 0 and 100.

    \item \textbf{Intraclass Correlation Coefficient} ($\mathrm{ICC}$) \cite{icc}, a standard metric in inter-rater reliability studies that assesses the degree of absolute agreement between two measurement methods. Similar to $\mathrm{QWK}$, it typically ranges from 0--1, with scores closer to 1 indicating stronger consistency and absolute agreement between raters. We use $\mathrm{ICC}(3,1)$, a two-way mixed-effects model with absolute agreement, which treats the LLM and human annotators as fixed raters evaluating the same set of information units.
\end{itemize}

Table~\ref{tab:coverage_faithfulness_human_alignment} shows that our automated coverage evaluation achieves strong alignment with human annotations across all metrics. The $\mathrm{PA}@0$ of 75.0\% indicates that three-quarters of LLM judgments exactly match the averaged human scores, while $\mathrm{PA}@25$ of 93.2\% shows that the vast majority of automated assessments fall within one rating level of human judgment. The high $\mathrm{QWK}$ score of 0.919 and $\mathrm{ICC}$ of 0.926 both indicate substantial agreement, with values approaching 1.0 suggesting near-perfect consistency between automated and human evaluations. These results validate the reliability of our LLM-based coverage assessment system.

\begin{table*}[t]
\centering
\scriptsize
\caption{Alignment between LLM judge and human for coverage and faithfulness evaluation.}
\label{tab:coverage_faithfulness_human_alignment}
    \begin{tabular}{lcccc}
    \toprule
    \textbf{Evaluation} & $\mathrm{PA}@0$ & $\mathrm{PA}@25$ & $\mathrm{QWK}$ & $\mathrm{ICC}$ \\
    \midrule
    Coverage     & 75.0 & 93.2 & 0.919 & 0.926 \\
    Faithfulness & 85.2 & 97.4 & 0.663 & 0.664 \\
    \bottomrule
    \end{tabular}
\end{table*}

\subsubsection{Faithfulness}
\label{sec:human-alignment-faithfulness}

For faithfulness evaluation, we used the same annotation methodology and metrics as described in the coverage section. We use GPT-5 with high reasoning effort for both claim extraction and faithfulness validation. For each of the 10 sampled data points, we randomly selected 50 statements that had been extracted and evaluated by the LLM judge from model-generated reports, resulting in 500 statements in total. Each statement was accompanied by the LLM judge's score indicating the degree to which the statement was faithfully supported by the grounding documents.

Two independent expert annotators evaluated each statement's faithfulness using the same three-point scale: 0 for not faithfully supported by grounding documents, 50 for partially supported, and 100 for fully supported. The ratings from both annotators were averaged to obtain final human scores with five discrete levels (0, 25, 50, 75, 100), which were then compared with continuous LLM-generated scores using $\mathrm{PA}@0$, $\mathrm{PA}@25$, $\mathrm{QWK}$, and $\mathrm{ICC}$.

Table~\ref{tab:coverage_faithfulness_human_alignment} shows that faithfulness evaluation achieves moderate-to-good alignment between automated and human assessments. The $\mathrm{PA}@0$ of 85.2\% indicates frequent exact agreement, while the $\mathrm{PA}@25$ of 97.4\% shows that automated judgments nearly always fall within one rating level of human assessments. However, the $\mathrm{QWK}$ of 0.663 and $\mathrm{ICC}$ of 0.664 are lower than those for coverage, suggesting moderate agreement. This pattern indicates that while the LLM judge's faithfulness assessments are practically aligned with human judgment in terms of ordinal ranking (as evidenced by high $\mathrm{PA}$ metrics), determining the precise degree of faithful support is inherently more nuanced and subjective than coverage assessment, making perfect agreement more challenging.

\section{Experimental Setup}

We evaluate five state-of-the-art LLMs spanning proprietary and open-source families. Because our grounding documents include images (Academic categories) and long contexts, we restrict selection to natively multimodal models with at least a 128K-token context window. The two proprietary models are GPT-5~\cite{gpt5} (400K context, ``high'' reasoning effort) and Claude-4.5-Sonnet~\cite{claude4.5} (200K context). The three open-source MoE/dense vision-language models are Qwen3-VL-30B-A3B~\cite{qwen-vl, qwen2-vl, qwen2.5-vl, qwen3} (30B total / 3.3B active, 256K context), Phi-4-multimodal~\cite{phi4} (5.6B dense, 128K context), and Kimi-VL-A3B~\cite{kimi-vl} (16B / 2.8B active, 128K context). All five models receive the same one-shot generation prompt, conditioned on the sample report $S$ and the grounding documents $G$.

\section{Results}
\label{sec:results}

\begin{table*}[t]
\centering
\scriptsize
\caption{Combined Evaluation Results across Structural, Coverage, and Faithfulness Metrics.}
\setlength{\tabcolsep}{3pt}
\begin{tabularx}{\linewidth}{cLCCCCC}
    \toprule
    \textbf{Metric} & \textbf{Data} & \textbf{GPT-5} & \textbf{Claude-4.5-Sonnet} & \textbf{Qwen3-VL-A3B} & \textbf{Phi-4-multi} & \textbf{Kimi-VL-A3B} \\
    \midrule
    \multirow{5}{*}{\rotatebox[origin=c]{90}{\textbf{Structural}}}
        & \cellcolor[gray]{.9} \textbf{Overall} & \cellcolor[gray]{.9} 0.298 & \cellcolor[gray]{.9} 0.314 & \cellcolor[gray]{.9} \textbf{0.315} & \cellcolor[gray]{.9} 0.253 & \cellcolor[gray]{.9} 0.204 \\
        & Financial Ann. & 0.331 & \textbf{0.392} & 0.390 & 0.241 & 0.177 \\
        & Financial Qtr. & \textbf{0.414} & 0.406 & 0.413 & 0.361 & 0.317 \\
        & Academic Comp. & 0.201 & \textbf{0.206} & 0.205 & 0.190 & 0.165 \\
        & Academic Simp. & 0.248 & \textbf{0.251} & 0.250 & 0.222 & 0.158 \\
    \midrule
    \multirow{5}{*}{\rotatebox[origin=c]{90}{\textbf{Coverage}}}
        & \cellcolor[gray]{.9} \textbf{Overall} & \cellcolor[gray]{.9} \textbf{0.779} & \cellcolor[gray]{.9} 0.770 & \cellcolor[gray]{.9} 0.548 & \cellcolor[gray]{.9} 0.258 & \cellcolor[gray]{.9} 0.298 \\
        & Financial Ann. & 0.764 & \textbf{0.860} & 0.748 & 0.403 & 0.436 \\
        & Financial Qtr. & 0.856 & \textbf{0.889} & 0.766 & 0.457 & 0.478 \\
        & Academic Comp. & \textbf{0.707} & 0.611 & 0.373 & 0.088 & 0.090 \\
        & Academic Simp. & \textbf{0.787} & 0.720 & 0.306 & 0.085 & 0.186 \\
    \midrule
    \multirow{5}{*}{\rotatebox[origin=c]{90}{\textbf{Faithfulness}}}
        & \cellcolor[gray]{.9} \textbf{Overall} & \cellcolor[gray]{.9} \textbf{0.914} & \cellcolor[gray]{.9} 0.843 & \cellcolor[gray]{.9} 0.545 & \cellcolor[gray]{.9} 0.361 & \cellcolor[gray]{.9} 0.625 \\
        & Financial Ann. & \textbf{0.913} & 0.844 & 0.698 & 0.591 & 0.807 \\
        & Financial Qtr. & \textbf{0.931} & 0.910 & 0.712 & 0.469 & 0.657 \\
        & Academic Comp. & \textbf{0.902} & 0.796 & 0.458 & 0.216 & 0.277 \\
        & Academic Simp. & \textbf{0.908} & 0.822 & 0.314 & 0.169 & 0.758 \\
    \bottomrule
\end{tabularx}
\label{tab:combined_results}
\end{table*}


\begin{figure*}[!t]
\centering

\begin{minipage}[t]{0.495\linewidth}
\begin{tcolorbox}[exstyle, equal height group=cs1, title={Gold Report}]
\begin{lstlisting}[style=mdstyle,escapeinside={(*}{*)}]
# # Q1 FY2026 Consolidated Financial and Operating Review

## ## 1. Executive Summary and Strategic Context
Salesforce delivered solid first-quarter results with revenue of approximately $9.8 billion, up 8% year over year. Income from operations was about $1.9 billion ... Diluted EPS were $1.59 ...

## ## 2. Financial Performance Overview
Total revenue grew to $9.829 billion, and gross profit increased to $7.564 billion ... Income from operations increased 13.6% to $1.942 billion ... Restructuring charges of $36 million ...

## ## 3. Revenue Composition and Customer Metrics
...
## ## 7. Capital Returns: Share Repurchases and Dividends
... 10 million shares ... for approximately $2.681 billion at an average price of $273.42 per share.
...
## ## 14. Subsequent Events and Other Disclosures
...
\end{lstlisting}
\end{tcolorbox}
\end{minipage}%
\hfill%
\begin{minipage}[t]{0.495\linewidth}
\begin{tcolorbox}[exstyle, equal height group=cs1, title={GPT-5 (Str: 0.421, Cov: 0.910, Fth: 0.985)}]
\begin{lstlisting}[style=mdstyle,escapeinside={(*}{*)}]
# # Q1 FY2026 Consolidated Financial and Operating Review

## ## 1. Executive Summary and Strategic Context
Total revenue was $9.83 billion for the three months ended April 30, 2025, up 8 percent year over year. Income from operations increased to $1.94 billion ... Diluted EPS was $1.59 ...

## ## 2. Financial Performance Overview
Gross profit increased to $7.56 billion (77 percent of revenue) ... Income from operations rose 14 percent in the quarter to $1.94 billion ... Restructuring expense was $36 million ...

## ## 3. Revenue Composition and Customer Metrics
...
## ## 7. Capital Returns: Share Repurchases and Dividends
... repurchases in the quarter totaling approximately (*\textcolor{blue}{10 million shares}*) for approximately (*\textcolor{blue}{\$2.68 billion}*).
...
## ## 14. Subsequent Events and Other Disclosures
... (*\textcolor{blue}{at an average price of \$273.42 per share}*) ...
\end{lstlisting}
\end{tcolorbox}
\end{minipage}

\vspace{4pt}

\begin{minipage}[t]{0.495\linewidth}
\begin{tcolorbox}[exstyle, equal height group=cs2, title={Claude-4.5-Sonnet (Str: 0.416, Cov: 0.880, Fth: 0.955)}]
\begin{lstlisting}[style=mdstyle,escapeinside={(*}{*)}]
# # Q1 FY2026 Consolidated Financial and Operating Review

## ## 1. Executive Summary and Strategic Context
Total revenue was $9.83 billion for the three months ended April 30, 2025, up 8 percent year over year. Income from operations increased to $1.94 billion ... Diluted EPS was $1.59 ...

## ## 2. Financial Performance Overview
Gross profit increased to $7.56 billion (77.0 percent of revenue) ... Income from operations rose 14 percent in the quarter to $1.94 billion ... Restructuring expense was $36 million ...

## ## 3. Revenue Composition and Customer Metrics
...
## ## 7. Capital Returns: Share Repurchases and Dividends
... repurchases during the quarter, totaling approximately (*\textcolor{blue}{10 million shares}*) for approximately (*\textcolor{blue}{\$2.7 billion}*) 
...
| Total Q1 FY2026 | ... | (*\textcolor{blue}{\$273.42}*) | ...
## ## 14. Subsequent Events and Other Disclosures
\end{lstlisting}
\end{tcolorbox}
\end{minipage}%
\hfill%
\begin{minipage}[t]{0.495\linewidth}
\begin{tcolorbox}[exstyle, equal height group=cs2, title={Qwen3-VL-A3B (Str: 0.409, Cov: 0.796, Fth: 0.635)}]
\begin{lstlisting}[style=mdstyle,escapeinside={(*}{*)}]
# # (*\textcolor{red}{Q3 FY2025}*) Consolidated Financial and Operating Review

## ## 1. Executive Summary and Strategic Context
Total revenue was $9.83 billion for the three months ended April 30, 2025 ...

## ## 2. Financial Performance Overview
Total revenue grew 8 percent in the quarter and 9 percent year-to-date. Gross profit increased to $7.56 billion ... Income from operations rose 14 percent in the quarter to $1.94 billion and (*\textcolor{red}{59 percent}*) year-to-date to (*\textcolor{red}{\$5.62 billion}*).

## ## 3. Revenue Composition and Customer Metrics
...
## ## 7. Capital Returns: Share Repurchases and Dividends
... totaling approximately (*\textcolor{blue}{10 million shares}*) for approximately (*\textcolor{blue}{\$2.7 billion}*) year-to-date 

(*\textcolor{red}{(average repurchase price not reported)}*)
...
## ## 14. Subsequent Events and Other Disclosures
\end{lstlisting}
\end{tcolorbox}
\end{minipage}

\vspace{4pt}

\begin{minipage}[t]{0.495\linewidth}
\begin{tcolorbox}[exstyle, equal height group=cs3, title={Phi-4-multimodal (Str: 0.429, Cov: 0.505, Fth: 0.305)}]
\begin{lstlisting}[style=mdstyle,escapeinside={(*}{*)}]
# # (*\textcolor{red}{Q3 FY2025}*) Consolidated Financial and Operating Review

## ## 1. Executive Summary and Strategic Context
Total revenue was (*\textcolor{red}{\$9.44 billion}*) for the three months ended (*\textcolor{red}{October 31, 2024}*) ... Diluted EPS was (*\textcolor{red}{\$1.58}*) ...

## ## 2. Financial Performance Overview
Gross profit increased to (*\textcolor{red}{\$7.34 billion (77.7\%)}*) ... Income from operations rose (*\textcolor{red}{26 percent}*) ... Restructuring expense was (*\textcolor{red}{\$56 million}*) ...

## ## 3. Revenue Composition and Customer Metrics
...
## ## 7. Capital Returns: Share Repurchases and Dividends
... totaling approximately (*\textcolor{red}{30 million shares}*) for approximately (*\textcolor{red}{\$7.7 billion}*) year-to-date 

(*\textcolor{red}{(average repurchase price not reported)}*)
...
## ## 14. Subsequent Events and Other Disclosures
...
\end{lstlisting}
\end{tcolorbox}
\end{minipage}%
\hfill%
\begin{minipage}[t]{0.495\linewidth}
\begin{tcolorbox}[exstyle, equal height group=cs3, title={Kimi-VL-A3B (Str: 0.430, Cov: 0.725, Fth: 0.665)}]
\begin{lstlisting}[style=mdstyle,escapeinside={(*}{*)}]
# # (*\textcolor{red}{Q4 FY2025}*) Consolidated Financial and Operating Review

## ## 1. Executive Summary and Strategic Context
Total revenue was $9.8 billion for the three months ended April 30, 2025, up 8% year-over-year ... Diluted EPS was $1.59 ...

## ## 2. Financial Performance Overview
Total revenue grew 8 percent in the quarter and 9 percent year-to-date ... Income from operations rose (*\textcolor{red}{20\%}*) ... Diluted EPS up (*\textcolor{red}{26\%}*) in the quarter ...

## ## 3. Revenue Composition and Customer Metrics
...
## ## 7. Capital Returns: Share Repurchases and Dividends
... totaling approximately (*\textcolor{blue}{10 million shares}*) for approximately (*\textcolor{blue}{\$2.7 billion}*) year-to-date 

(*\textcolor{red}{(average repurchase price not reported)}*)
...
## ## 14. Subsequent Events and Other Disclosures
...
\end{lstlisting}
\end{tcolorbox}
\end{minipage}

\caption{%
  Case study: Salesforce 2026 Q1. All models achieve similar structural similarity (0.409--0.430). \textcolor{red}{Red} marks unfaithful or uncovered content; \textcolor{blue}{blue} marks correctly covered or faithful content.%
}
\label{fig:case-study-reports}
\end{figure*}

\begin{figure*}[t]
    \centering
    \begin{subfigure}[t]{0.32\linewidth}
        \centering
        \includegraphics[width=\linewidth]{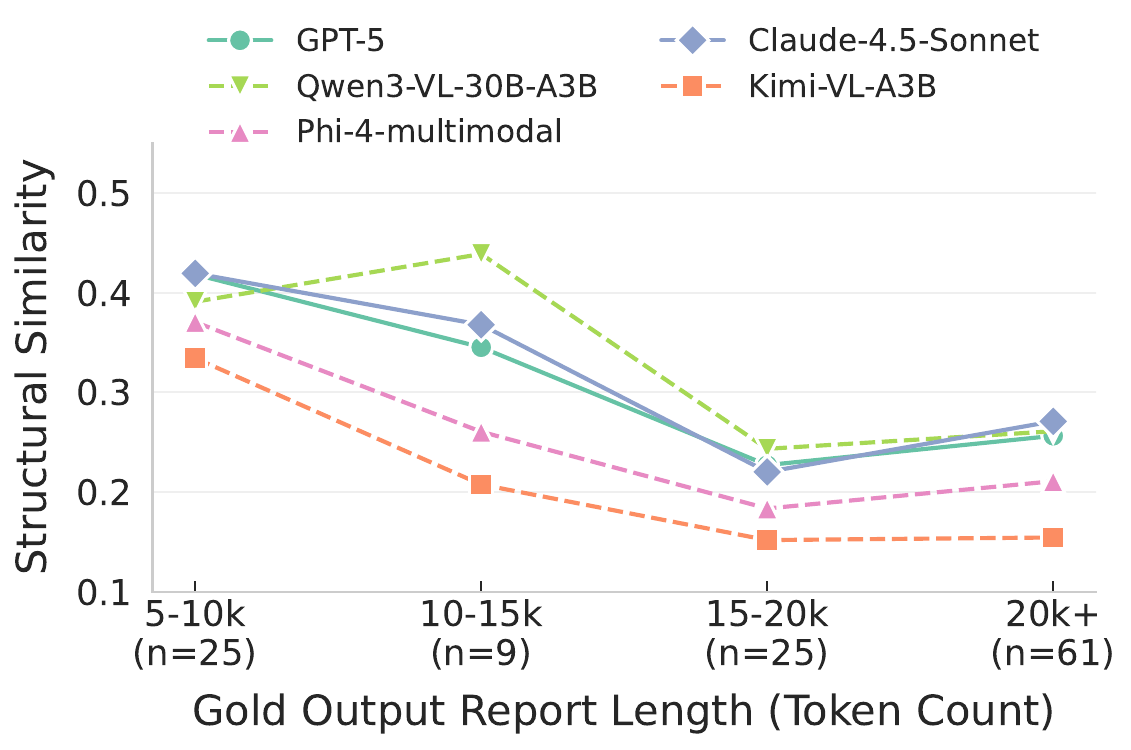}
    \end{subfigure}\hfill
    \begin{subfigure}[t]{0.32\linewidth}
        \centering
        \includegraphics[width=\linewidth]{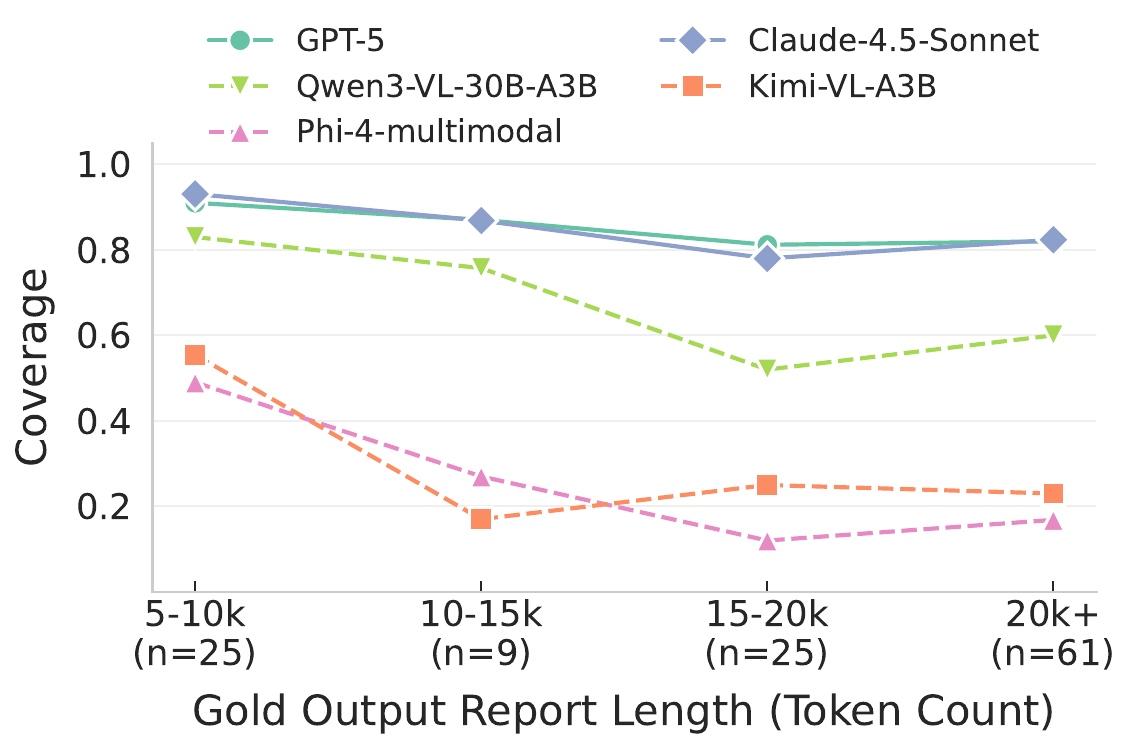}
    \end{subfigure}\hfill
    \begin{subfigure}[t]{0.32\linewidth}
        \centering
        \includegraphics[width=\linewidth]{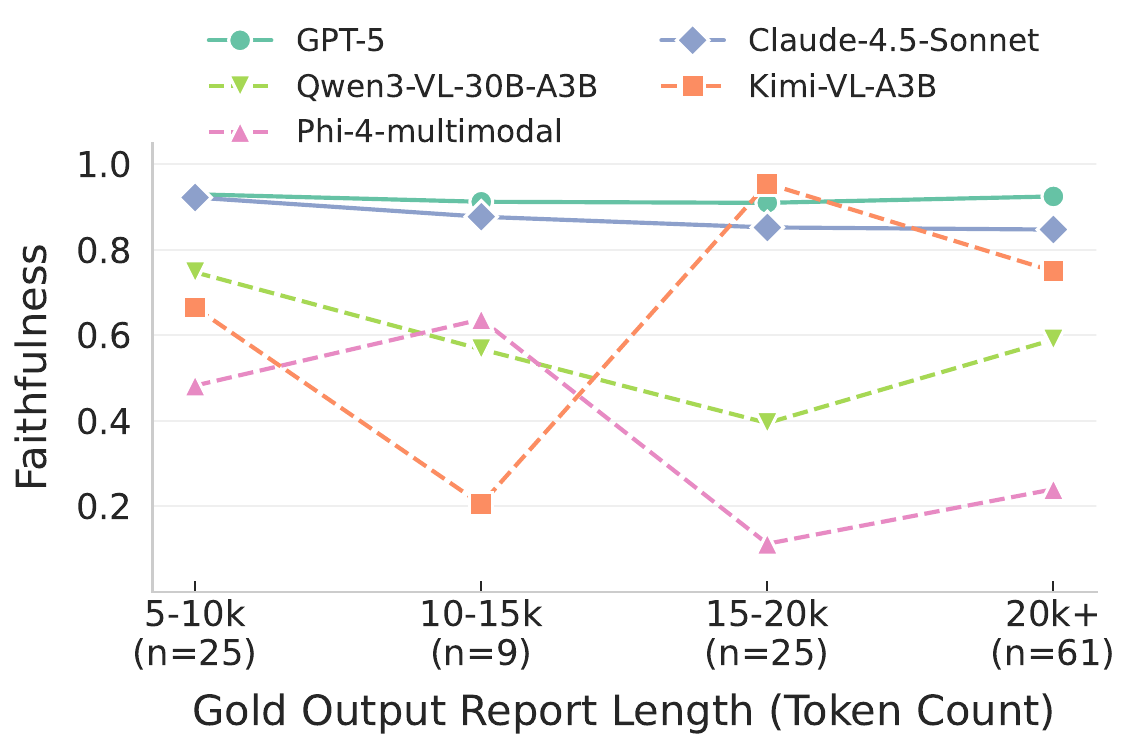}
    \end{subfigure}
    \caption{Performance trends versus gold output report length.}
    \label{fig:score_trend_vs_length}
\end{figure*}

\begin{figure*}[t]
    \centering
    \begin{subfigure}[t]{0.32\linewidth}
        \centering
        \includegraphics[width=\linewidth]{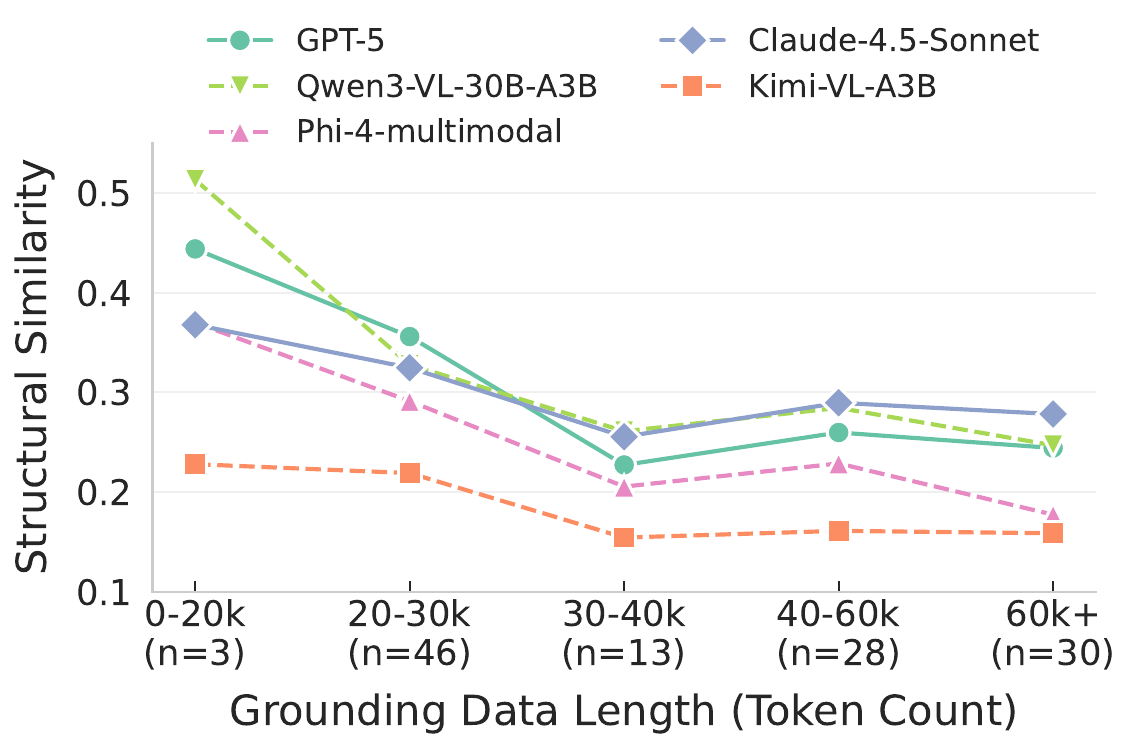}
    \end{subfigure}\hfill
    \begin{subfigure}[t]{0.32\linewidth}
        \centering
        \includegraphics[width=\linewidth]{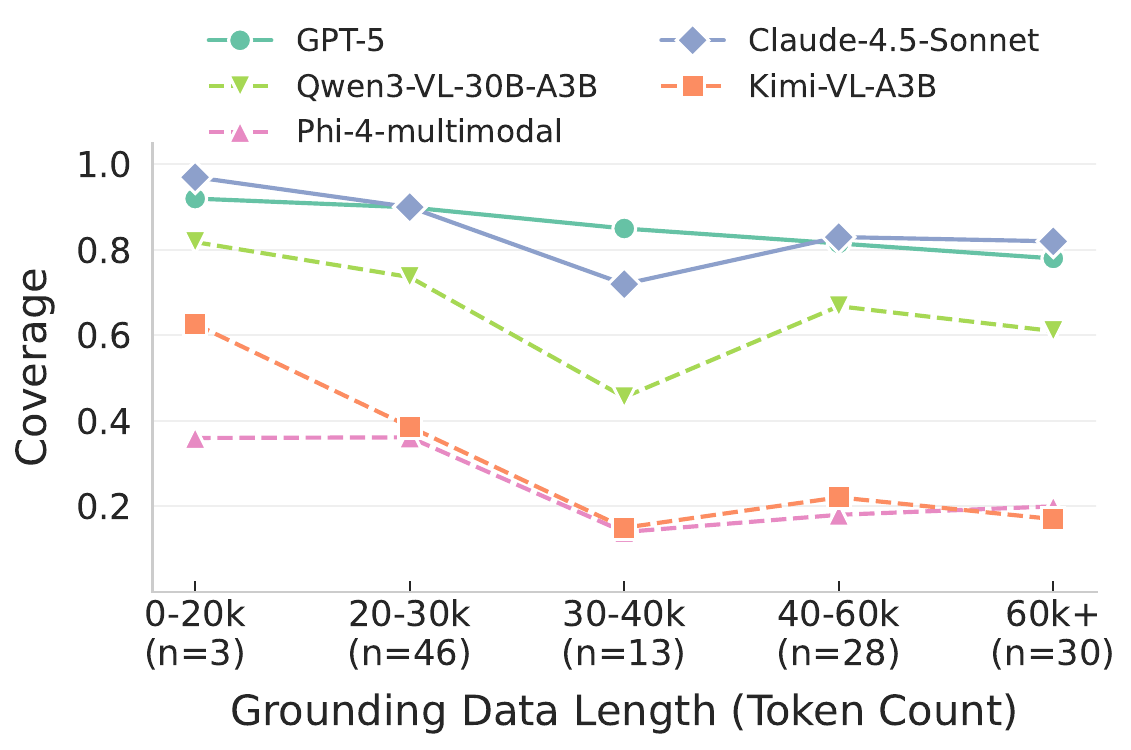}
    \end{subfigure}\hfill
    \begin{subfigure}[t]{0.32\linewidth}
        \centering
        \includegraphics[width=\linewidth]{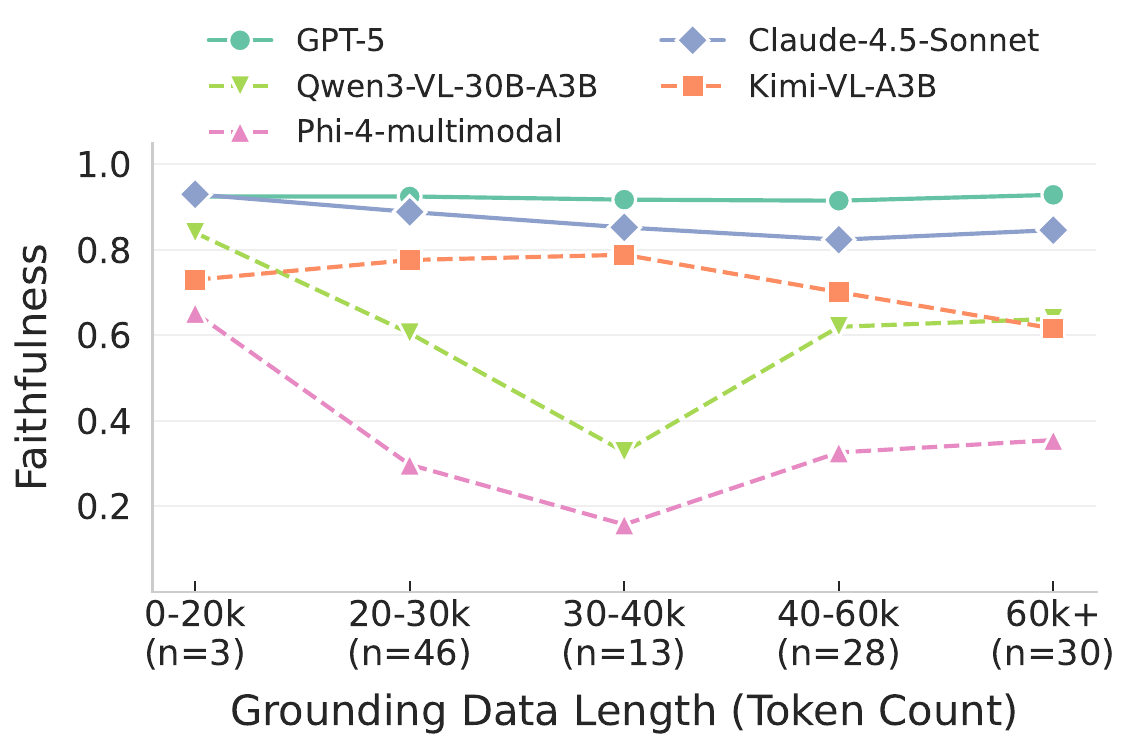}
    \end{subfigure}
    \caption{Performance trends versus grounding documents length.}
    \label{fig:score_trend_vs_grounding_length}
\end{figure*}

\begin{figure*}[t]
    \centering
    \begin{subfigure}[t]{0.32\linewidth}
        \centering
        \includegraphics[width=\linewidth]{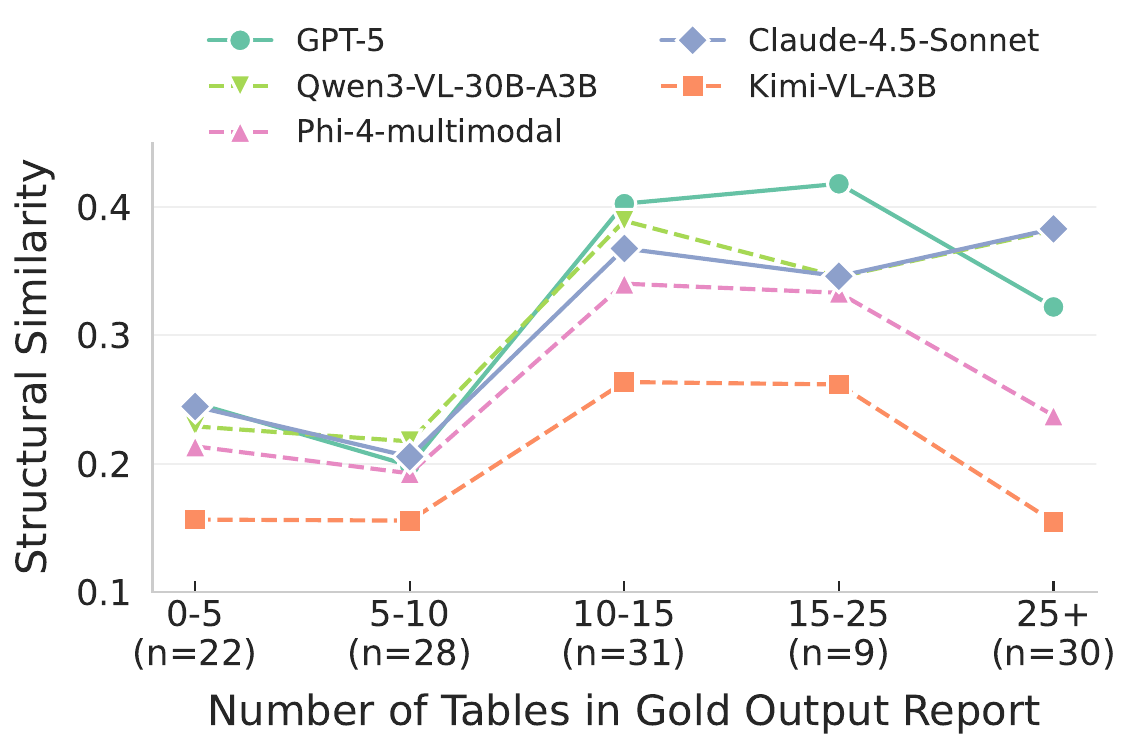}
    \end{subfigure}\hfill
    \begin{subfigure}[t]{0.32\linewidth}
        \centering
        \includegraphics[width=\linewidth]{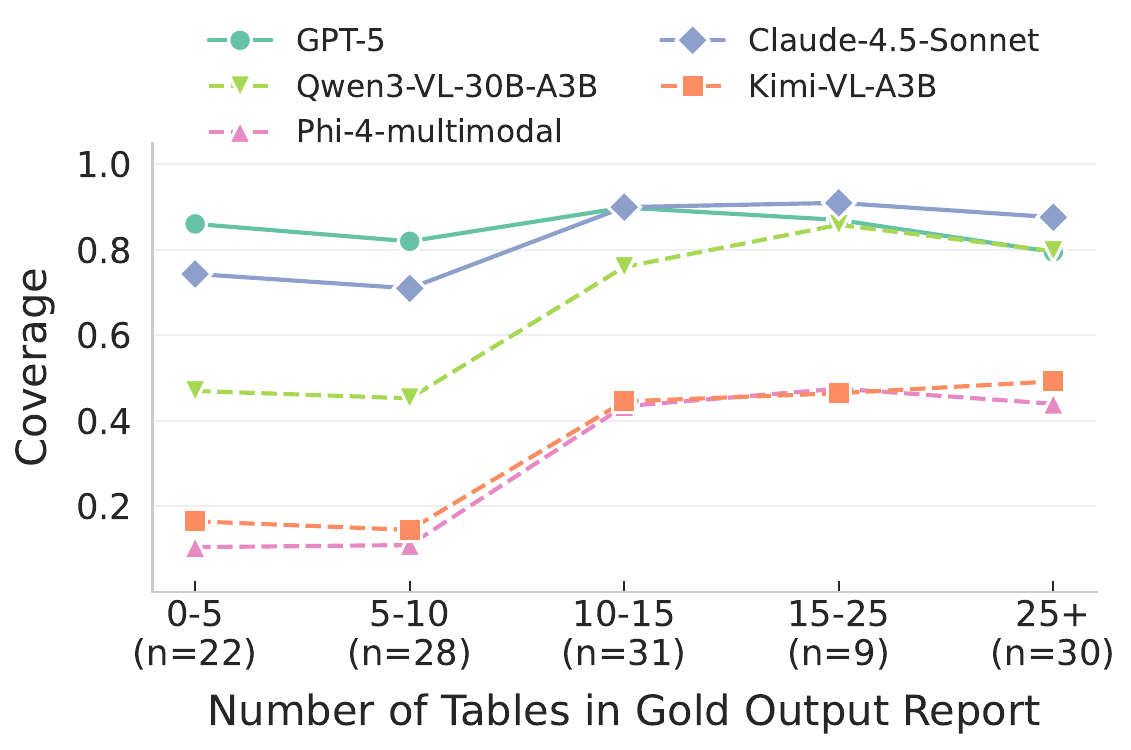}
    \end{subfigure}\hfill
    \begin{subfigure}[t]{0.32\linewidth}
        \centering
        \includegraphics[width=\linewidth]{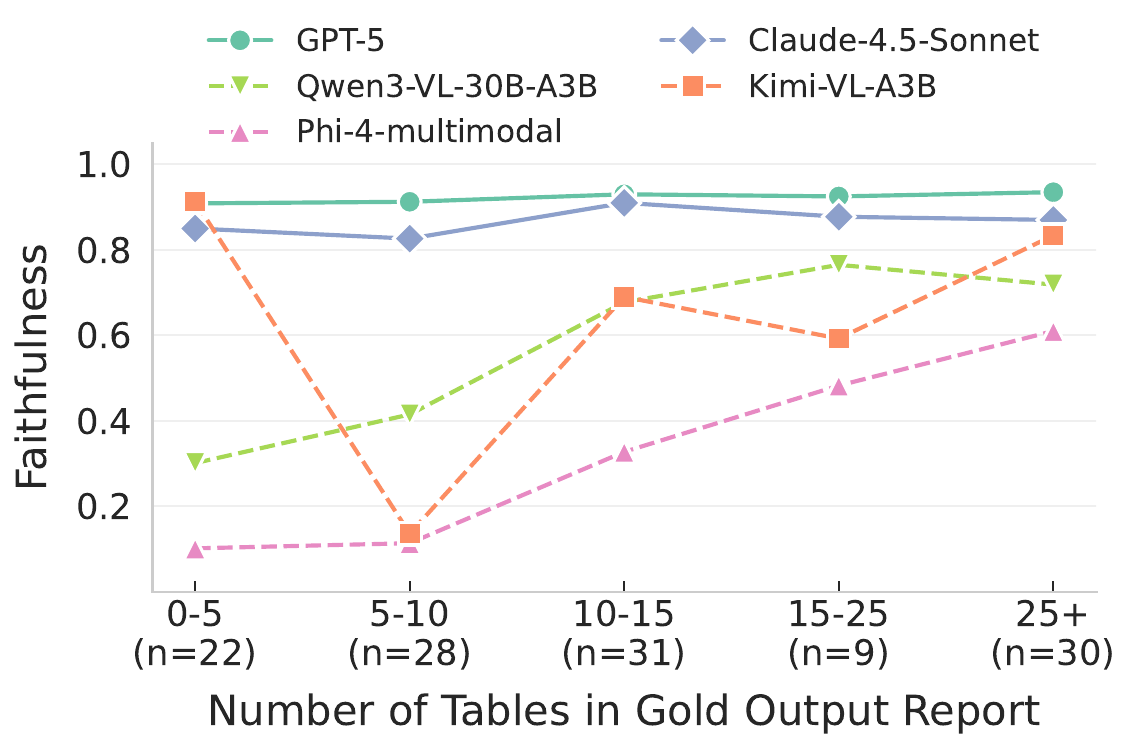}
    \end{subfigure}
    \caption{Performance trends versus the number of tables in gold output reports.}
    \label{fig:score_trend_vs_num_tables}
\end{figure*}

\subsection{Overall Performance}
\label{sec:overall-performance}

Table~\ref{tab:combined_results} reports the five LLMs across the three metrics. We highlight four findings.

\finding{Open-source models match proprietary models on structural similarity} The structural scores are tightly clustered across model families, suggesting that current open-source systems can often infer and reproduce the sample report's form.

\finding{Proprietary models significantly outperform open-source models on coverage and faithfulness} The largest gaps are not in layout replication but in content completeness and evidence grounding. This indicates that form transfer is easier than selecting all salient information and supporting each generated claim.

\finding{Models perform better on financial reports than on academic reports} Financial reports have standardized, numerical formats that make recurring content and evidence easier to identify. Academic reports are more flexible and multimodal, producing larger drops, especially for open-source models.

\finding{Model performance generally degrades with increased task complexity} Harder categories reduce structural similarity and coverage for most models. Faithfulness is more stable for proprietary systems but remains high-variance for weaker models, particularly on academic data.

\subsection{Case Study}
\label{sec:case-study}

Figure~\ref{fig:case-study-reports} illustrates the patterns of \S\ref{sec:overall-performance} on a concrete Financial Quarterly instance (Salesforce 2026 Q1). All five models reproduce the gold report's section layout with nearly identical structural similarity (0.409--0.430), confirming that template replication is the easiest sub-task. However, the generated content diverges sharply. GPT-5 and Claude correctly identify the reporting period as Q1~FY2026 and faithfully reproduce key figures such as revenue, growth rates, and the average share repurchase price of \$273.42. In contrast, open-source models misidentify the quarter (Qwen3 and Phi-4 label it ``Q3~FY2025''; Kimi labels it ``Q4~FY2025''), hallucinate incorrect figures (e.g., Phi-4 reports revenue of \$9.44B instead of \$9.83B and operations growth of 26\% instead of 14\%), and omit specific details such as the repurchase price that all three proprietary models cover. This case concretely demonstrates how structural parity can coexist with large gaps in coverage and faithfulness.

\subsection{Performance Trend Analysis}
In order to better understand the factors driving model performance variation, we analyze how performance changes with key characteristics, including gold output report length, grounding document length, and the number of tables in the gold output report.

\paragraph{Length of Gold Output Report.}
Figure~\ref{fig:score_trend_vs_length} shows three consistent patterns across length buckets. First, structural similarity generally declines as reports grow longer, indicating that matching the target template becomes harder for long outputs. Second, coverage decreases for all models from the shortest bucket onward, with only minor rebounds at the longest bucket for some. Third, GPT-5 is the only model with clearly stable faithfulness across length buckets; Claude-4.5-Sonnet shows a slight decline, and open-source models fluctuate substantially, consistent with weaker grounding robustness. \finding{Longer outputs mainly hurt structural quality and coverage, while faithfulness remains stable only for GPT-5}

\paragraph{Length of Grounding Context.}
Figure~\ref{fig:score_trend_vs_grounding_length} shows that all three metrics decline as grounding contexts grow longer, with some intermediate non-monotonicity. Structural similarity and coverage drop for all models. On faithfulness, only GPT-5 stays stable; Claude-4.5-Sonnet drifts slightly, and open-source models degrade with higher variance. \finding{Longer grounding contexts make evidence selection and reasoning harder, generally reduce structural quality and coverage, and increase faithfulness instability for open-source models}

\paragraph{Number of Tables in Gold Output Report.}
Figure~\ref{fig:score_trend_vs_num_tables} reveals several trends with table count. Structural similarity peaks at a moderate-to-high number of tables rather than at very low counts. Coverage rises with table count for Claude-4.5-Sonnet and all three open-source models, suggesting tabular content is easier to cover. Faithfulness for open-source models also trends upward despite local fluctuations, suggesting better grounding with more tabular evidence. \finding{Reports with more tabular content tend to improve coverage and faithfulness, especially for open-source models}

\section{Conclusion}

We introduced \textsc{Frame}, a 120-instance benchmark for form-restricted report generation, where models must transfer a sample report's structure while grounding new content. We also developed three human-validated metrics for evaluating this task, covering structure, information coverage, and faithfulness.

Across five LLMs, our results show that template replication is substantially easier than complete and faithful content grounding: open-source models approach proprietary models on structural similarity but lag on coverage and faithfulness. Performance is stronger on financial reports than academic reports, declines with longer and more complex inputs, and improves when evidence is more table-rich. These findings highlight form-restricted report generation as a practical but still challenging test of long-form, evidence-grounded generation.

\section*{Limitations}
\label{sec:limitations}

\textsc{Frame} has two potential limitations about the scope of the dataset and the evaluation metrics.

\paragraph{Domain coverage.} The benchmark covers two domains, financial filings (SEC 10-Q/10-K) and arXiv literature reviews, chosen for their structured form and public availability. Other professional report-writing settings (e.g., clinical reports, legal briefs, technical documentation) likely exhibit different structural conventions and grounding patterns, and our results may not directly transfer.

\paragraph{Reliance on an LLM judge.} Coverage and faithfulness are scored by an LLM judge (GPT-5 at high reasoning effort). We validate the judge against expert human annotations and report alignment metrics (PA, QWK, ICC) in \S\ref{sec:human-alignment}, but the judge inherits the underlying model's biases and may underperform on content outside its training distribution. The structural metric does not depend on an LLM judge but does rely on a text embedding model to score node similarity.

\bibliography{main}

\begin{thebibliography}{36}
\providecommand{\natexlab}[1]{#1}

\bibitem[{Abdin et~al.(2024)Abdin, Aneja, Behl, Bubeck, Eldan, Gunasekar,
  Harrison, Hewett, Javaheripi, Kauffmann, Lee, Lee, Li, Liu, Mendes, Nguyen,
  Price, Rosa, Saarikivi, Salim, Shah, Wang, Ward, Wu, Yu, Zhang, and
  Zhang}]{phi4}
Marah Abdin, Jyoti Aneja, Harkirat Behl, S{\'e}bastien Bubeck, Ronen Eldan,
  Suriya Gunasekar, Michael Harrison, Russell~J. Hewett, Mojan Javaheripi,
  Piero Kauffmann, James~R. Lee, Yin~Tat Lee, Yuanzhi Li, Weishung Liu, Caio
  C.~T. Mendes, Anh Nguyen, Eric Price, Gustavo~de Rosa, Olli Saarikivi, and 8
  others. 2024.
\newblock Phi-4 technical report.
\newblock \emph{arXiv preprint arXiv:2412.08905}.

\bibitem[{Anthropic(2025)}]{claude4.5}
Anthropic. 2025.
\newblock \href {https://www.anthropic.com/claude-sonnet-4-5-system-card}
  {Claude sonnet 4.5 system card}.
\newblock \emph{Anthropic Technical Report}.

\bibitem[{Bai et~al.(2023)Bai, Bai, Yang, Wang, Tan, Wang, Lin, Zhou, and
  Zhou}]{qwen-vl}
Jinze Bai, Shuai Bai, Shusheng Yang, Shijie Wang, Sinan Tan, Peng Wang, Junyang
  Lin, Chang Zhou, and Jingren Zhou. 2023.
\newblock Qwen-vl: A versatile vision-language model for understanding,
  localization, text reading, and beyond.
\newblock \emph{arXiv preprint arXiv:2308.12966}.

\bibitem[{Bai et~al.(2025)Bai, Chen, Liu, Wang, Ge, Song, Dang, Wang, Wang,
  Tang, Zhong, Zhu, Yang, Li, Wan, Wang, Ding, Fu, Xu, Ye, Zhang, Xie, Cheng,
  Zhang, Yang, Xu, and Lin}]{qwen2.5-vl}
Shuai Bai, Keqin Chen, Xuejing Liu, Jialin Wang, Wenbin Ge, Sibo Song, Kai
  Dang, Peng Wang, Shijie Wang, Jun Tang, Humen Zhong, Yuanzhi Zhu, Mingkun
  Yang, Zhaohai Li, Jianqiang Wan, Pengfei Wang, Wei Ding, Zheren Fu, Yiheng
  Xu, and 8 others. 2025.
\newblock Qwen2.5-vl technical report.
\newblock \emph{arXiv preprint arXiv:2502.13923}.

\bibitem[{Cohen(1968)}]{qwk}
Jacob Cohen. 1968.
\newblock \href {https://doi.org/10.1037/h0026256} {Weighted kappa: Nominal
  scale agreement provision for scaled disagreement or partial credit.}
\newblock \emph{Psychological Bulletin}, 70(4):213--220.

\bibitem[{Durmus et~al.(2020)Durmus, He, and Diab}]{durmus-etal-2020-feqa}
Esin Durmus, He~He, and Mona Diab. 2020.
\newblock \href {https://doi.org/10.18653/v1/2020.acl-main.454} {{FEQA}: A
  question answering evaluation framework for faithfulness assessment in
  abstractive summarization}.
\newblock In \emph{Proceedings of the 58th Annual Meeting of the Association
  for Computational Linguistics}, pages 5055--5070, Online. Association for
  Computational Linguistics.

\bibitem[{Fabbri et~al.(2019)Fabbri, Li, She, Li, and
  Radev}]{fabbri2019multinews}
Alexander Fabbri, Irene Li, Tianwei She, Suyi Li, and Dragomir Radev. 2019.
\newblock {Multi-News}: A large-scale multi-document summarization dataset and
  abstractive hierarchical model.
\newblock In \emph{Proceedings of the 57th Annual Meeting of the Association
  for Computational Linguistics}, pages 1074--1084.

\bibitem[{Grattafiori et~al.(2024)}]{dubey2024llama3}
Aaron Grattafiori and 1 others. 2024.
\newblock The {Llama} 3 herd of models.
\newblock \emph{arXiv preprint arXiv:2407.21783}.

\bibitem[{Hunt(1986)}]{pa}
Ronald~J Hunt. 1986.
\newblock \href {https://doi.org/10.1177/00220345860650020701} {Percent
  agreement, pearson's correlation, and kappa as measures of inter-examiner
  reliability}.
\newblock \emph{Journal of Dental Research}, 65(2):128--130.

\bibitem[{Islam et~al.(2023)Islam, Kannappan, Kiela, Qian, Scherber, and
  Vidgen}]{islam2023financebench}
Pranab Islam, Anand Kannappan, Douwe Kiela, Rebecca Qian, Nino Scherber, and
  Bertie Vidgen. 2023.
\newblock {FinanceBench}: A new benchmark for financial question answering.
\newblock \emph{arXiv preprint arXiv:2311.11944}.

\bibitem[{K{\"o}ksal et~al.(2023)K{\"o}ksal, Schick, Korhonen, and
  Sch{\"u}tze}]{koksal2023longform}
Abdullatif K{\"o}ksal, Timo Schick, Anna Korhonen, and Hinrich Sch{\"u}tze.
  2023.
\newblock {LongForm}: Effective instruction tuning with reverse instructions.
\newblock \emph{arXiv preprint arXiv:2304.08460}.

\bibitem[{Kulkarni et~al.(2025)Kulkarni, Zhang, Moniz, Ge, Tseng,
  Piraviperumal, Swayamdipta, and Yu}]{kulkarni-etal-2025-evaluating}
Atharva Kulkarni, Yuan Zhang, Joel Ruben~Antony Moniz, Xiou Ge, Bo-Hsiang
  Tseng, Dhivya Piraviperumal, Swabha Swayamdipta, and Hong Yu. 2025.
\newblock \href {https://doi.org/10.18653/v1/2025.findings-emnlp.1035}
  {Evaluating evaluation metrics {--} the mirage of hallucination detection}.
\newblock In \emph{Findings of the Association for Computational Linguistics:
  EMNLP 2025}, pages 19013--19032, Suzhou, China. Association for Computational
  Linguistics.

\bibitem[{Lewis et~al.(2021)Lewis, Perez, Piktus, Petroni, Karpukhin, Goyal,
  Küttler, Lewis, tau Yih, Rocktäschel, Riedel, and
  Kiela}]{lewis2021retrievalaugmentedgenerationknowledgeintensivenlp}
Patrick Lewis, Ethan Perez, Aleksandra Piktus, Fabio Petroni, Vladimir
  Karpukhin, Naman Goyal, Heinrich Küttler, Mike Lewis, Wen tau Yih, Tim
  Rocktäschel, Sebastian Riedel, and Douwe Kiela. 2021.
\newblock \href {https://arxiv.org/abs/2005.11401} {Retrieval-augmented
  generation for knowledge-intensive nlp tasks}.
\newblock \emph{Preprint}, arXiv:2005.11401.

\bibitem[{Liu et~al.(2023)Liu, Iter, Xu, Wang, Xu, and Zhu}]{liu-etal-2023-g}
Yang Liu, Dan Iter, Yichong Xu, Shuohang Wang, Ruochen Xu, and Chenguang Zhu.
  2023.
\newblock \href {https://doi.org/10.18653/v1/2023.emnlp-main.153} {{G}-eval:
  {NLG} evaluation using gpt-4 with better human alignment}.
\newblock In \emph{Proceedings of the 2023 Conference on Empirical Methods in
  Natural Language Processing}, pages 2511--2522, Singapore. Association for
  Computational Linguistics.

\bibitem[{Liu et~al.(2025)Liu, Zhu, Liu, Chen, Liu, Guo, Zhang, Bao, and
  Luo}]{liu2025generalizedtreeeditdistance}
Yuntian Liu, Tao Zhu, Xiaoyang Liu, Yu~Chen, Zhaoxuan Liu, Qingfeng Guo,
  Jiashuo Zhang, Kangjie Bao, and Tao Luo. 2025.
\newblock \href {https://arxiv.org/abs/2507.07399} {Generalized tree edit
  distance (gted): A faithful evaluation metric for statement
  autoformalization}.
\newblock \emph{Preprint}, arXiv:2507.07399.

\bibitem[{Min et~al.(2023)Min, Krishna, Lyu, Lewis, tau Yih, Koh, Iyyer,
  Zettlemoyer, and Hajishirzi}]{Min2023FactScore}
Sewon Min, Kalpesh Krishna, Xinxi Lyu, Mike Lewis, Wen tau Yih, Pang~Wei Koh,
  Mohit Iyyer, Luke Zettlemoyer, and Hannaneh Hajishirzi. 2023.
\newblock \href {https://doi.org/10.18653/v1/2023.emnlp-main.741}
  {{FA}ct{S}core: Fine-grained atomic evaluation of factual precision in long
  form text generation}.
\newblock In \emph{Proceedings of the 2023 Conference on Empirical Methods in
  Natural Language Processing}, pages 12076--12100. Association for
  Computational Linguistics.

\bibitem[{Nenkova and Passonneau(2004)}]{nenkova2004pyramid}
Ani Nenkova and Rebecca~J. Passonneau. 2004.
\newblock \href {https://aclanthology.org/N04-1019/} {Evaluating content
  selection in summarization: The pyramid method}.
\newblock In \emph{Proceedings of the Human Language Technology Conference of
  the North American Chapter of the Association for Computational Linguistics:
  HLT-NAACL 2004}, pages 145--152. Association for Computational Linguistics.

\bibitem[{Ning et~al.(2024)Ning, Lin, Zhou, Wang, Yang, and
  Wang}]{ning2024skeletonofthoughtpromptingllmsefficient}
Xuefei Ning, Zinan Lin, Zixuan Zhou, Zifu Wang, Huazhong Yang, and Yu~Wang.
  2024.
\newblock \href {https://arxiv.org/abs/2307.15337} {Skeleton-of-thought:
  Prompting llms for efficient parallel generation}.
\newblock \emph{Preprint}, arXiv:2307.15337.

\bibitem[{OpenAI(2023)}]{openai2023gpt4}
OpenAI. 2023.
\newblock {GPT-4} technical report.
\newblock \emph{arXiv preprint arXiv:2303.08774}.

\bibitem[{Paa{\ss}en(2018)}]{DBLP:journals/corr/abs-1805-06869}
Benjamin Paa{\ss}en. 2018.
\newblock \href {https://arxiv.org/abs/1805.06869} {Revisiting the tree edit
  distance and its backtracing: {A} tutorial}.
\newblock \emph{CoRR}, abs/1805.06869.

\bibitem[{Que et~al.(2024)Que, Duan, He, Mou, Zhou, Liu, Rong, Wang, Yang,
  Zhang, Peng, Zhang, Zhang, and Chen}]{que2024hellobenchevaluatinglongtext}
Haoran Que, Feiyu Duan, Liqun He, Yutao Mou, Wangchunshu Zhou, Jiaheng Liu,
  Wenge Rong, Zekun~Moore Wang, Jian Yang, Ge~Zhang, Junran Peng, Zhaoxiang
  Zhang, Songyang Zhang, and Kai Chen. 2024.
\newblock \href {https://arxiv.org/abs/2409.16191} {Hellobench: Evaluating long
  text generation capabilities of large language models}.
\newblock \emph{Preprint}, arXiv:2409.16191.

\bibitem[{Shao et~al.(2019)Shao, Huang, Wen, Xu, and Zhu}]{shao-etal-2019-long}
Zhihong Shao, Minlie Huang, Jiangtao Wen, Wenfei Xu, and Xiaoyan Zhu. 2019.
\newblock \href {https://doi.org/10.18653/v1/D19-1321} {Long and diverse text
  generation with planning-based hierarchical variational model}.
\newblock In \emph{Proceedings of the 2019 Conference on Empirical Methods in
  Natural Language Processing and the 9th International Joint Conference on
  Natural Language Processing (EMNLP-IJCNLP)}, pages 3257--3268, Hong Kong,
  China. Association for Computational Linguistics.

\bibitem[{Shen et~al.(2025)Shen, Wang, Mishra, Xu, Teng, and
  Ding}]{shen-etal-2025-slot}
Zhengyuan Shen, Darren Yow-Bang Wang, Soumya~Smruti Mishra, Zhichao Xu, Yifei
  Teng, and Haibo Ding. 2025.
\newblock \href {https://doi.org/10.18653/v1/2025.emnlp-industry.32} {{SLOT}:
  Structuring the output of large language models}.
\newblock In \emph{Proceedings of the 2025 Conference on Empirical Methods in
  Natural Language Processing: Industry Track}, pages 472--491, Suzhou (China).
  Association for Computational Linguistics.

\bibitem[{Shrout and Fleiss(1979)}]{icc}
Patrick~E Shrout and Joseph~L Fleiss. 1979.
\newblock \href {https://doi.org/10.1037/0033-2909.86.2.420} {Intraclass
  correlations: uses in assessing rater reliability.}
\newblock \emph{Psychological Bulletin}, 86(2):420--428.

\bibitem[{Singh et~al.(2025)}]{gpt5}
Aaditya Singh and 1 others. 2025.
\newblock \href {https://arxiv.org/abs/2601.03267} {Openai gpt-5 system card}.
\newblock \emph{arXiv preprint arXiv:2601.03267}.

\bibitem[{Team et~al.(2025)Team, Du, Yin, Xing, Qu, Wang, Chen, Zhang, Du, Wei,
  Wang, Zhang, Du, Wang, Yuan, Lu, Li, Sung, Wei, Lai, Zhu, Ding, Hu, Yang,
  Zhang, Wu, Yao, Lu, Wang, Gao, Zheng, Li, Su, Wang, Deng, Qiu, Xie, Wang,
  Liu, Yan, Ouyang, Chen, Sui, Yu, Dong, Dong, Xu, Cheng, Gu, Zhou, Liu, Cao,
  Yu, Song, Bai, Song, He, Huang, Xu, Yuan, Yao, Wu, Zu, Zhou, Wang, Charles,
  Zhong, Li, Hu, Chen, Wang, Liu, Miao, Qin, Chen, Bao, Wang, Kang, Liu, Du,
  Wu, Wang, Yan, Zhou, Li, Jiang, Zhang, Yang, Huang, Huang, Zhao, and
  Chen}]{kimi-vl}
Kimi Team, Angang Du, Bohong Yin, Bowei Xing, Bowen Qu, Bowen Wang, Cheng Chen,
  Chenlin Zhang, Chenzhuang Du, Chu Wei, Congcong Wang, Dehao Zhang, Dikang Du,
  Dongliang Wang, Enming Yuan, Enzhe Lu, Fang Li, Flood Sung, Guangda Wei, and
  73 others. 2025.
\newblock \href {https://arxiv.org/abs/2504.07491} {{Kimi-VL} technical
  report}.
\newblock \emph{Preprint}, arXiv:2504.07491.

\bibitem[{Wang et~al.(2024{\natexlab{a}})Wang, Bai, Tan, Wang, Fan, Bai, Chen,
  Liu, Wang, Ge, Fan, Dang, Du, Ren, Men, Liu, Zhou, Zhou, and Lin}]{qwen2-vl}
Peng Wang, Shuai Bai, Sinan Tan, Shijie Wang, Zhihao Fan, Jinze Bai, Keqin
  Chen, Xuejing Liu, Jialin Wang, Wenbin Ge, Yang Fan, Kai Dang, Mengfei Du,
  Xuancheng Ren, Rui Men, Dayiheng Liu, Chang Zhou, Jingren Zhou, and Junyang
  Lin. 2024{\natexlab{a}}.
\newblock Qwen2-vl: Enhancing vision-language model's perception of the world
  at any resolution.
\newblock \emph{arXiv preprint arXiv:2409.12191}.

\bibitem[{Wang et~al.(2024{\natexlab{b}})Wang, Guo, Yao, Zhang, Zhang, Wu,
  Zhang, Dai, Zhang, Wen, Ye, Zhang, and Zhang}]{wang2024autosurvey}
Yidong Wang, Qi~Guo, Wenjin Yao, Hongbo Zhang, Xin Zhang, Zhen Wu, Meishan
  Zhang, Xinyu Dai, Min Zhang, Qingsong Wen, Wei Ye, Shikun Zhang, and Yue
  Zhang. 2024{\natexlab{b}}.
\newblock {AutoSurvey}: Large language models can automatically write surveys.
\newblock \emph{arXiv preprint arXiv:2406.10252}.

\bibitem[{Wiseman et~al.(2017)Wiseman, Shieber, and
  Rush}]{wiseman2017challenges}
Sam Wiseman, Stuart Shieber, and Alexander Rush. 2017.
\newblock Challenges in data-to-document generation.
\newblock In \emph{Proceedings of the 2017 Conference on Empirical Methods in
  Natural Language Processing}, pages 2253--2263.

\bibitem[{Wu et~al.(2023)Wu, Irsoy, Lu, Dabravolski, Dredze, Gehrmann,
  Kambadur, Rosenberg, and Mann}]{wu2023bloomberggpt}
Shijie Wu, Ozan Irsoy, Steven Lu, Vadim Dabravolski, Mark Dredze, Sebastian
  Gehrmann, Prabhanjan Kambadur, David Rosenberg, and Gideon Mann. 2023.
\newblock {BloombergGPT}: A large language model for finance.
\newblock \emph{arXiv preprint arXiv:2303.17564}.

\bibitem[{Wu and Deng(2025)}]{wu2025implementinglongtextstyle}
Yusen Wu and Xiaotie Deng. 2025.
\newblock \href {https://arxiv.org/abs/2505.07888} {Implementing long text
  style transfer with llms through dual-layered sentence and paragraph
  structure extraction and mapping}.
\newblock \emph{Preprint}, arXiv:2505.07888.

\bibitem[{Yang et~al.(2025)}]{qwen3}
An~Yang and 1 others. 2025.
\newblock \href {https://arxiv.org/abs/2505.09388} {Qwen3 technical report}.

\bibitem[{Zha et~al.(2023)Zha, Yang, Li, and Hu}]{zha2023alignscore}
Yuheng Zha, Yichi Yang, Ruichen Li, and Zhiting Hu. 2023.
\newblock \href {https://doi.org/10.18653/v1/2023.acl-long.634}
  {{A}lign{S}core: Evaluating factual consistency with a unified alignment
  function}.
\newblock In \emph{Proceedings of the 61st Annual Meeting of the Association
  for Computational Linguistics (Volume 1: Long Papers)}, pages 11328--11348,
  Toronto, Canada. Association for Computational Linguistics.

\bibitem[{Zhang et~al.(2024)Zhang, Wan, Cattan, Klein, and
  Dagan}]{zhang2024qapyramid}
Shiyue Zhang, David Wan, Arie Cattan, Ayal Klein, and Ido Dagan. 2024.
\newblock \href {https://arxiv.org/abs/2412.07096} {{QAPyramid}: Fine-grained
  evaluation of content selection for text summarization}.
\newblock \emph{arXiv preprint arXiv:2412.07096}.

\bibitem[{Zhang et~al.(2019)Zhang, Kishore, Wu, Weinberger, and
  Artzi}]{zhang2019bertscore}
Tianyi Zhang, Varsha Kishore, Felix Wu, Kilian~Q. Weinberger, and Yoav Artzi.
  2019.
\newblock \href {https://arxiv.org/abs/1904.09675} {Bertscore: Evaluating text
  generation with bert}.
\newblock \emph{arXiv preprint arXiv:1904.09675}.

\bibitem[{Zhao et~al.(2023)Zhao, Zhou, Li, Tang, Wang, Hou, Min, Zhang, Zhang,
  Dong et~al.}]{zhao2023survey}
Wayne~Xin Zhao, Kun Zhou, Junyi Li, Tianyi Tang, Xiaolei Wang, Yupeng Hou,
  Yingqian Min, Beichen Zhang, Junjie Zhang, Zican Dong, and 1 others. 2023.
\newblock A survey of large language models.
\newblock \emph{arXiv preprint arXiv:2303.18223}.

\end{thebibliography}


\appendix

\section{Benchmark Construction Examples}
\label{app:benchmark-examples}

This appendix provides illustrative examples for the data-synthesis pipeline (\S\ref{ssec:data-synthesis}) and a concrete benchmark instance, complementing the high-level description in the main text.


\begin{figure*}[!t]
\centering

\begin{minipage}[t]{0.495\linewidth}

\begin{tcolorbox}[exstyle, equal height group=syn1, title={(1)~~Raw iXBRL Filing}]
\begin{lstlisting}[style=xmlstyle]
<ix:header>
  <ix:resources>
    <xbrli:context id="FI2025Q1">
      <xbrli:period>
        <xbrli:instant>2025-02-28</xbrli:instant>
      </xbrli:period>
    </xbrli:context>
    <xbrli:context id="FI2025Q1YTD">
      <xbrli:period>
        <xbrli:startDate>2024-12-01</xbrli:startDate>
        <xbrli:endDate>2025-02-28</xbrli:endDate>
      </xbrli:period>
    </xbrli:context>
    <xbrli:unit id="USD">
      <xbrli:measure>iso4217:USD</xbrli:measure>
    </xbrli:unit>
    <!-- ... more context ... -->
  </ix:resources>
</ix:header>

<ix:nonFraction name="us-gaap:Revenues"
  contextRef="FI2025Q1" unitRef="USD"
  decimals="-6">5,714</ix:nonFraction>

<ix:nonFraction name="us-gaap:NetIncomeLoss"
  contextRef="FI2025Q1" unitRef="USD"
  decimals="-6">1,811</ix:nonFraction>

<!-- ~4 MB: XBRL namespaces, inline styles,
     scripts, HTML tables mixed with prose -->
\end{lstlisting}
\end{tcolorbox}

\vspace{-5pt}

\begin{tcolorbox}[exstyle, equal height group=syn2,
  title={(3)~~Synthesized Report Structure}]
\begin{lstlisting}[style=jsonstyle]
{
  "heading": "Adobe Inc. Q1 FY2025 Report",
  "children": [
    {
      "heading": "1. Executive Summary",
      "children": [
        {
          "type": "text",
          "objective": "Summarize registrant context, revenue mix, segment/geo highlights, profitability, capital allocation, and liquidity.",
          "questions": [
            "What were total revenue and the subscription/product/services mix?",
            "Which segments/regions led growth?",
            "What capital actions occurred?",
            ...
          ]
        }
      ]
    },
    {
      "heading": "2. Business Context",
      "children": [
        {
          "type": "table",
          "columns": [
            "Segment", "Revenue", "Gross Profit", "Margin"
          ],
          "objective": "Show segment revenue and margins."
        },
        ...
      ]
    },
...
\end{lstlisting}
\end{tcolorbox}

\end{minipage}%
\hfill%
\begin{minipage}[t]{0.495\linewidth}

\begin{tcolorbox}[exstyle, equal height group=syn1,
  title={(2)~~Grounding Documents $D$}]
\begin{lstlisting}[style=filestyle]
----- 10 grounding documents in total -----
revenue_segments_geography_contract.md
equity_earnings_per_share_and_aoci.md
condensed_statements_of_income.md
condensed_statements_comprehensive.md
condensed_statements_stockholders.md
...

----- revenue_segments_geography_contract.md -----
# # Revenue, Segments, Geography, and Contract Balances

Adobe provides products, services and solutions used to imagine, create, manage, deliver, measure, optimize and engage with content across media and surfaces...

## ## Segment Results
| Metric          | Dig.Media | Dig.Exp | Total |
|-----------------|----------:|--------:|------:|
| Revenue ($M)    |     4,227 |   1,414 | 5,714 |
| Gross profit    |     4,028 |   1,012 | 5,092 |
| Gross margin    |       95% |     72% |   89% |
...

----- equity_earnings_per_share_and_aoci.md -----
# # Equity, Earnings Per Share, and Accumulated Other Comprehensive Income

Reclassification adjustments for gains and losses on available-for-sale securities are recorded in other income (expense), net ...

\end{lstlisting}
\end{tcolorbox}

\vspace{-5pt}

\begin{tcolorbox}[exstyle, equal height group=syn2,
  title={(4)~~Gold Report $R$}]
\begin{lstlisting}[style=mdstyle]
# # Adobe Inc. Q1 FY2025 Analytical Report

## ## 1. Executive Summary
Adobe Inc. (NASDAQ: ADBE) reported total revenue of $5,714M for Q1 FY2025 (ended Feb 28, 2025), led by subscription revenue of $5,483M (96%).

Segment performance: Digital Media $4,227M, Digital Experience $1,414M, Publishing $73M. Geographic: Americas 59.6%, EMEA 26.3%, APAC 14.1%. All regions grew 7-14% YoY.

Profitability: gross profit $5,092M (89% margin), operating income $2,163M, net income $1,811M, diluted EPS $4.14 (vs. $1.36 prior year). Effective tax rate 17% (vs. 36%), aided by absence of prior-year $1B termination fee.

Capital: issued $2.0B new senior notes (4.75-5.30% due 2028--2035); repaid $1.5B; repurchased $3.25B shares (7.0M shares). Liquidity: $6,758M cash + $677M short-term investments, undrawn $1.5B revolver.

## ## 2. Business Context

| Segment      | Revenue | Gross Profit | Margin |
|--------------|--------:|-------------:|-------:|
| Digital Med. |  $4,227 |       $4,028 |    95% |
| Digital Exp. |  $1,414 |       $1,012 |    72% |
| Total        |  $5,714 |       $5,092 |    89% |

| Region   | Feb 2025 | YoY   | Share  |
|----------|---------:|------:|-------:|
| Americas |   $3,405 |  +9%  | 59.6%  |
| EMEA     |   $1,502 | +14%  | 26.3%  |
| APAC     |     $807 |  +7%  | 14.1%  |
...
\end{lstlisting}
\end{tcolorbox}

\end{minipage}

\caption{%
  Synthesis pipeline for a single $(D, R)$ pair (Adobe 2025 Q1 10-Q).
  (1)~The raw iXBRL filing is split into semantically coherent grounding documents~(2).
  An LLM synthesizes a hierarchical JSON report structure~(3), which guides
  generation of the gold report~(4).%
}
\label{fig:synthesis-example}
\end{figure*}

Figure~\ref{fig:synthesis-example} walks through the four stages of the synthesis pipeline on Adobe's Q1~FY2025 10-Q. \textbf{(1)~Raw iXBRL Filing} shows a fragment of the raw EDGAR source: an HTML/XBRL document with namespace declarations, inline \code{<ix:nonFraction>} tags carrying numerical facts, and prose interleaved with HTML tables (~4~MB total). \textbf{(2)~Grounding Documents $D$} shows the result of the file-split and content-generation stages: the LLM partitions the filing into 10 semantically coherent Markdown documents (e.g., \code{revenue\_segments\_geography\_contract.md}, \code{condensed\_statements\_of\_income.md}), each with clean prose and well-formatted tables. \textbf{(3)~Synthesized Report Structure} shows the hierarchical JSON outline, with each leaf carrying an \code{objective} and guiding \code{questions} for the final-stage generator. \textbf{(4)~Gold Report $R$} shows the resulting Markdown report, produced by filling each placeholder while conditioning on the grounding set, the structure, and the per-section questions.

%

\begin{figure*}[!t]
\centering
\begin{minipage}{\linewidth}

\begin{tcolorbox}[exstyle, title={Sample Report $S$ (from Adobe 2025 Q1)}]
\begin{lstlisting}[style=mdstyle]
# # Adobe Inc. Q1 FY2025 Consolidated Analytical Report

## ## 1. Executive Summary and Key Highlights
Adobe Inc. (NASDAQ: ADBE) reported total revenue of $5,714M for Q1 FY2025 (ended Feb 28, 2025), led by subscription revenue of $5,483M (96%). Segment performance was led by Digital Media ($4,227M) and Digital Experience ($1,414M). Profitability: gross profit $5,092M (89% margin), operating income $2,163M, net income $1,811M, diluted EPS $4.14.

## ## 2. Business Context and Segment Performance

| Segment        | Revenue | Gross Profit | Margin |
|----------------|--------:|-------------:|-------:|
| Digital Media  |  $4,227 |       $4,028 |    95% |
| Digital Exp.   |  $1,414 |       $1,012 |    72% |
| Pub. & Adv.    |     $73 |          $52 |    71% |

## ## 3. Revenue Mix and Contract Dynamics
Subscription revenue accounted for 96% of total revenue ($5,483 million of $5,714 million) in Q1 FY2025, with product and services representing the remaining 4%. Subscription revenue is generally recognized ratably over the contract term as control of services transfers...

## ## 4. Financial Statements Overview
Revenue growth in Q1 FY2025 was driven by subscription across Digital Media and Digital Experience, yielding consolidated gross profit of $5,092 million on total cost of revenue of $622 million...

## ## 5. Liquidity, Investments, and Market Risk...
\end{lstlisting}
\end{tcolorbox}

\vspace{-5pt}

\begin{tcolorbox}[exstyle, title={Grounding Documents $G$ (from Adobe 2025 Q2)}]
\begin{lstlisting}[style=filestyle]
----- 10 grounding documents in total -----
revenue_and_segment_disclosures_q2_2025.md        condensed_financial_statements_q2_2025.md
debt_and_liquidity_q2_2025.md                     capital_structure_and_per_share_data.md
...

----- revenue_and_segment_disclosures_q2_2025.md -----
# # Revenue, Segments, and Customer Mix (Q2 FY2025)

## ## Operating Segments
| Segment           | Revenue | Cost of Rev | Gross Profit | Margin |
|-------------------|--------:|------------:|-------------:|-------:|
| Digital Media     |  $4,345 |        $208 |       $4,137 |    95% |
| Digital Experience|  $1,457 |        $408 |       $1,049 |    72% |
| Pub. & Adv.       |     $71 |         $22 |          $49 |    69% |

## ## Geographic Revenue
| Region   | Q2 2025 | Q2 2024 | YoY Growth |
|----------|--------:|--------:|-----------:|
| Americas |  $3,500 |  $3,188 |       +10% |
| EMEA     |  $1,541 |  $1,361 |       +13% |
| APAC     |    $832 |    $760 |        +9% |

For the quarter, segment growth was driven by subscription momentum across offerings. Digital Media revenue rose to $4.35 billion, up 11% year over year, led by strength in Creative Cloud flagship apps and Acrobat ...
\end{lstlisting}
\end{tcolorbox}

\vspace{-5pt}

\begin{tcolorbox}[exstyle, title={Gold Report $Y$ (from Adobe 2025 Q2)}]
\begin{lstlisting}[style=mdstyle]
# # Adobe Inc. Q2 FY2025 Consolidated Analytical Report

## ## 1. Executive Summary and Key Highlights
Adobe Inc. (NASDAQ: ADBE) reported total revenue of $5,873M for Q2 FY2025 (ended May 30, 2025), led by subscription revenue of $5,641M (96%). Digital Media $4,345M, up 11% YoY; Digital Experience $1,457M, up 10% YoY. Gross profit $5,235M, operating income $2,109M, net income $1,691M, diluted EPS $3.94...

## ## 2. Business Context and Segment Performance

| Segment        | Revenue | Gross Profit | Margin |
|----------------|--------:|-------------:|-------:|
| Digital Media  |  $4,345 |       $4,137 |    95% |
| Digital Exp.   |  $1,457 |       $1,049 |    72% |
| Pub. & Adv.    |     $71 |          $49 |    69% |

## ## 3. Revenue Mix and Contract Dynamics
Subscription revenue comprised 96% of total revenue in Q2 FY2025, reflecting the ratable recognition of performance obligations as services are delivered. Product revenue represented 2%, and services and other revenue accounted for 2%...

## ## 4. Financial Statements Overview
Revenue growth and gross margin in Q2 FY2025 were driven by subscription momentum across Digital Media and Digital Experience, with gross profit rising to $5,235 million. Cost of revenue increased to $638 million, reflecting higher hosting and data center expenses and AI inferencing costs...

## ## 5. Liquidity, Investments, and Market Risk...
\end{lstlisting}
\end{tcolorbox}

\end{minipage}

\caption{%
  A benchmark instance $(S, G, Y)$ from Adobe's quarterly filings.
  $S$ defines the sample structure, $G$ supplies updated grounding evidence, and $Y$ is the gold report that preserves the form of~$S$ with the content of~$G$. At inference time the model receives $(S, G)$ and must produce~$Y$.%
}
\label{fig:benchmark-example}
\end{figure*}

Figure~\ref{fig:benchmark-example} shows a concrete benchmark instance from the temporally ordered Adobe snapshots (\S\ref{ssec:task-formulation}): \textbf{Sample Report $S$} is the gold report from 2025~Q1, \textbf{Grounding Documents $G$} are the 2025~Q2 grounding set, and \textbf{Gold Report $Y$} is the gold report from those Q2 grounding documents. At inference time the model receives $(S, G)$ and must produce a report that follows $S$'s form while drawing content from $G$; $Y$ is the reference target.

\section{Coverage and Faithfulness Examples}
\label{app:coverage-faithfulness-examples}


\begin{figure*}[!t]
\centering

\begin{minipage}[t]{0.495\linewidth}

\begin{tcolorbox}[exstyle, equal height group=cov, title={Extracted Information Units $\mathcal{U}_{\text{ref}}$}]
\begin{lstlisting}[style=jsonstyle]
[
  ...
  {
    "unit_id": 7,
    "unit_type": "entities",
    "content": "San Jose, California",
    "source_sentence": "The companys principal executive offices are in San Jose, California."
  },
  ...
  {
    "unit_id": 26,
    "unit_type": "statistics",
    "content": "Gross profit was $5,235 million versus $4,711 million in the prior-year quarter.",
    "source_sentence": "Gross profit was $5,235 million versus $4,711 million in the prior-year quarter, operating income was $2,109 million versus ..."
  },
  ...
  {
    "unit_id": 31,
    "unit_type": "statistics",
    "content": "Weighted average diluted shares declined to 428.9 million from 451.4 million.",
    "source_sentence": ...
  },
  ...
  {
    "unit_id": 47,
    "unit_type": "statistics",
    "content": "December 2024 ASR settled for 6.3 million shares at an average price of $433.84.",
    "source_sentence": ...
  },
  ...
  {
    "unit_id": 81,
    "unit_type": "factual_claims",
    "content": "Adobe operates a subscription-based model anchored in three core cloud offerings.",
    "source_sentence": ...
  },
  ...    // 791 units total
]
\end{lstlisting}
\end{tcolorbox}

\end{minipage}%
\hfill%
\begin{minipage}[t]{0.495\linewidth}

\begin{tcolorbox}[exstyle, equal height group=cov, title={Coverage Evaluation Results}]
\begin{lstlisting}[style=jsonstyle]
[
  {
    "unit_id": 7,
    "content": "San Jose, California",
    "score": 0.0,
    "explanation": "The report does not mention San Jose, California."
  },
  {
    "unit_id": 26,
    "content": "Gross profit was $5,235 million versus $4,711 million in the prior-year quarter.",
    "score": 1.0,
    "explanation": "The report states gross profit of $5,235 million versus $4,711 million in the prior-year quarter."
  },
  {
    "unit_id": 31,
    "content": "Weighted average diluted shares declined to 428.9 million from 451.4 million.",
    "score": 0.95,
    "explanation": "Diluted shares declined to approximately 429 million from 451 million, closely matching."
  },
  {
    "unit_id": 47,
    "content": "December 2024 ASR settled for 6.3 million shares at an average price of $433.84.",
    "score": 0.0,
    "explanation": "The report does not mention the December 2024 ASR settlement details."
  },
  {
    "unit_id": 81,
    "content": "Adobe operates a subscription-based model anchored in three core cloud offerings.",
    "score": 1.0,
    "explanation": "The report describes Adobe's subscription-based model anchored in Creative Cloud, Document Cloud, and Experience Cloud."
  },
  ...    // IUR = 0.894
]
\end{lstlisting}
\end{tcolorbox}

\end{minipage}

\caption{%
  Coverage evaluation for the Adobe Q2~FY2025 instance. Information units are extracted from the gold report~(1), then each unit is checked against the GPT-5 generated report~(2) and assigned a coverage score.%
}
\label{fig:coverage-example}
\end{figure*}

Figure~\ref{fig:coverage-example} illustrates both stages of coverage evaluation on the Adobe Q2~FY2025 instance. The left panel shows a subset of the information units extracted from the gold report: each unit records its type (entity, statistic, factual claim, etc.), content, and the source sentence it was derived from. The right panel shows the corresponding coverage judgments against the GPT-5 generated report. Units receive a score of $1.0$ when the generated report faithfully reproduces the information (e.g., the gross profit comparison in unit~26), $0.95$ for near-matches with minor rounding differences (unit~31, where 428.9M is reported as approximately 429M), or $0.0$ when the generated report omits specific details such as the company's office location (unit~7) or a particular ASR settlement figure (unit~47).


\begin{figure*}[!t]
\centering

\begin{minipage}[t]{0.495\linewidth}

\begin{tcolorbox}[exstyle, equal height group=faith, title={Extracted Claims $\mathcal{C}$}]
\begin{lstlisting}[style=jsonstyle]
[
  ...
  {
    "claim_id": 7,
    "claim": "Subscription revenue represented 96% of total revenue for the quarter ended May 30, 2025.",
    "source_sentence": "For the quarter ended May 30, 2025, the company reported total revenue of $5,873 million, led by subscription revenue of $5,641 million (96% of total) ..."
  },
  ...
  {
    "claim_id": 26,
    "claim": "Gross profit was $5,235 million.",
    "source_sentence": "Gross profit was $5,235 million versus $4,711 million a year ago."
  },
  ...
  {
    "claim_id": 45,
    "claim": "Year to date, cash flow hedging gains were $39 million.",
    "source_sentence": "... partially offset by cash flow hedging gains of $9 million for the quarter (and $39 million year to date)."
  },
  ...
  {
    "claim_id": 72,
    "claim": "Creative Cloud includes Adobe Express",
    "source_sentence": "Creative Cloud includes flagship applications such as Photoshop, Illustrator, Lightroom, Premiere Pro and Adobe Stock, along with Adobe Express ..."
  },
  ...    // 788 claims total
]
\end{lstlisting}
\end{tcolorbox}

\end{minipage}%
\hfill%
\begin{minipage}[t]{0.495\linewidth}

\begin{tcolorbox}[exstyle, equal height group=faith, title={Faithfulness Evaluation Results}]
\begin{lstlisting}[style=jsonstyle]
[
  {
    "claim_id": 7,
    "claim": "Subscription revenue represented 96% of total revenue for the quarter ended May 30, 2025.",
    "support_level": 1.0,
    "explanation": "Subscription revenue was $5,641M and total revenue was $5,873M; 5,641/5,873 = 96.05%, i.e., 96%."
  },
  {
    "claim_id": 26,
    "claim": "Gross profit was $5,235 million.",
    "support_level": 0.75,
    "explanation": "Numerically correct for Q2 FY2025, but the claim omits the period, creating ambiguity since YTD gross profit is $10,327M."
  },
  {
    "claim_id": 45,
    "claim": "Year to date, cash flow hedging gains were $39 million.",
    "support_level": 1.0,
    "explanation": "Year to date, cash flow hedging gains were $39 million, per the revenue and segment disclosures."
  },
  {
    "claim_id": 72,
    "claim": "Creative Cloud includes Adobe Express",
    "support_level": 0.0,
    "explanation": "The filing lists Creative Cloud flagship apps (Photoshop, Lightroom, Illustrator) separately from Adobe Express; the claim is not supported."
  },
  ...    // Faithfulness = 0.88
]
\end{lstlisting}
\end{tcolorbox}

\end{minipage}

\caption{%
  Faithfulness evaluation for the Adobe Q2~FY2025 instance. Claims are extracted from the GPT-5 generated report (left), then each one of the sampled claim is verified against the grounding documents (right) and assigned a support level score.%
}
\label{fig:faithfulness-example}
\end{figure*}

Figure~\ref{fig:faithfulness-example} illustrates both stages of faithfulness verification on the Adobe Q2~FY2025 instance. The left panel shows a subset of the 788 atomic claims extracted from the GPT-5 generated report, each with the source sentence it was derived from. The right panel shows the verification results against the grounding documents. Claims receive a support score of $1.0$ when fully grounded (e.g., claim~7, where the subscription revenue percentage is arithmetically verified), $0.75$ when numerically correct but contextually ambiguous (claim~26, which omits the reporting period), or $0.0$ when the claim is not supported by the provided documents (claim~72, where the filing does not state that Creative Cloud includes Adobe Express).

\onecolumn
\section{Benchmark Construction Prompts}
\label{app:benchmark-prompts}


\providecolor{promptheading}{HTML}{0451A5}    
\providecolor{promptkey}{HTML}{0451A5}        
\providecolor{promptkeyword}{HTML}{1A6847}    
\providecolor{promptcomment}{HTML}{6A737D}    
\providecolor{promptstring}{HTML}{A31515}     

\lstdefinestyle{promptbase}{
  basicstyle=\ttfamily\fontsize{6pt}{7pt}\selectfont,
  breaklines=true,
  breakatwhitespace=false,
  breakindent=0pt,
  breakautoindent=false,
  columns=fixed,
  keepspaces=true,
  xleftmargin=0pt,
  frame=none,
  aboveskip=0pt,
  belowskip=0pt,
  showstringspaces=false,
  upquote=true,                       
}

\lstdefinestyle{promptmd}{
  style=promptbase,
  moredelim=[il][\color{promptheading}\bfseries]{\#\#\#\ },
  moredelim=[il][\color{promptheading}\bfseries]{\#\#\ },
  moredelim=[il][\color{promptheading}\bfseries]{\#\ },
  moredelim=[is][\bfseries]{**}{**},
  literate=
    *{"section\_type"}{{{\color{promptkey}"section\_type"}}}{14}
     {"heading"}{{{\color{promptkey}"heading"}}}{9}
     {"objective"}{{{\color{promptkey}"objective"}}}{11}
     {"children"}{{{\color{promptkey}"children"}}}{10}
     {"columns"}{{{\color{promptkey}"columns"}}}{9}
     {"file\_name"}{{{\color{promptkey}"file\_name"}}}{11}
     {"content"}{{{\color{promptkey}"content"}}}{9}
     {"files"}{{{\color{promptkey}"files"}}}{7}
     {"type"}{{{\color{promptkey}"type"}}}{6}
     {"questions"}{{{\color{promptkey}"questions"}}}{11}
     {"unit\_id"}{{{\color{promptkey}"unit\_id"}}}{9}
     {"unit\_type"}{{{\color{promptkey}"unit\_type"}}}{11}
     {"source\_sentence"}{{{\color{promptkey}"source\_sentence"}}}{17}
     {"confidence"}{{{\color{promptkey}"confidence"}}}{12}
     {"score"}{{{\color{promptkey}"score"}}}{7}
     {"explanation"}{{{\color{promptkey}"explanation"}}}{13}
     {"claim\_id"}{{{\color{promptkey}"claim\_id"}}}{10}
     {"claim"}{{{\color{promptkey}"claim"}}}{7}
     {"support\_level"}{{{\color{promptkey}"support\_level"}}}{15}
     {"accuracy\_status"}{{{\color{promptkey}"accuracy\_status"}}}{17}
     {"coverage\_score"}{{{\color{promptkey}"coverage\_score"}}}{16}
     {"error\_type"}{{{\color{promptkey}"error\_type"}}}{12}
     {"evidence\_found"}{{{\color{promptkey}"evidence\_found"}}}{16}
     {"is\_accurate"}{{{\color{promptkey}"is\_accurate"}}}{13}
     {"is\_covered"}{{{\color{promptkey}"is\_covered"}}}{12}
     {"severity"}{{{\color{promptkey}"severity"}}}{10}
     {"source\_excerpt"}{{{\color{promptkey}"source\_excerpt"}}}{16}
     {"sources"}{{{\color{promptkey}"sources"}}}{9}
     {"statement\_index"}{{{\color{promptkey}"statement\_index"}}}{17}
     {\{raw\_document\}}{{{\color{promptcomment}\itshape\{raw\_document\}}}}{14}
     {\{raw\_document\_top\_level\_section\}}{{{\color{promptcomment}\itshape\{raw\_document\_top\_level\_section\}}}}{36}
     {\{html\_paper\_example\}}{{{\color{promptcomment}\itshape\{html\_paper\_example\}}}}{20}
     {\{markdown\_paper\_structure\_example\}}{{{\color{promptcomment}\itshape\{markdown\_paper\_structure\_example\}}}}{34}
     {\{markdown\_grounding\_documents\_example\}}{{{\color{promptcomment}\itshape\{markdown\_grounding\_documents\_example\}}}}{36}
     {\{json\_output\_example\}}{{{\color{promptcomment}\itshape\{json\_output\_example\}}}}{21}
     {\{report\_structure\_template\}}{{{\color{promptcomment}\itshape\{report\_structure\_template\}}}}{25}
     {\{report\_template\}}{{{\color{promptcomment}\itshape\{report\_template\}}}}{16}
     {\{raw\_html\}}{{{\color{promptcomment}\itshape\{raw\_html\}}}}{10}
     {\{file\_name\}}{{{\color{promptcomment}\itshape\{file\_name\}}}}{11}
     {\{file\_content\}}{{{\color{promptcomment}\itshape\{file\_content\}}}}{14}
     {\{paper\_html\}}{{{\color{promptcomment}\itshape\{paper\_html\}}}}{12}
     {\{paper\_structure\}}{{{\color{promptcomment}\itshape\{paper\_structure\}}}}{17}
     {\{table\_html\}}{{{\color{promptcomment}\itshape\{table\_html\}}}}{12}
     {\{grounding\_documents\}}{{{\color{promptcomment}\itshape\{grounding\_documents\}}}}{20}
     {\{sample\_report\}}{{{\color{promptcomment}\itshape\{sample\_report\}}}}{14}
     {\{generated\_report\}}{{{\color{promptcomment}\itshape\{generated\_report\}}}}{17}
     {\{reference\_report\}}{{{\color{promptcomment}\itshape\{reference\_report\}}}}{17}
     {\{information\_units\}}{{{\color{promptcomment}\itshape\{information\_units\}}}}{19}
     {\{segment\_content\}}{{{\color{promptcomment}\itshape\{segment\_content\}}}}{17}
     {\{ground\_truth\}}{{{\color{promptcomment}\itshape\{ground\_truth\}}}}{14}
     {\{statements\}}{{{\color{promptcomment}\itshape\{statements\}}}}{12}
     {\{report\_structure\}}{{{\color{promptcomment}\itshape\{report\_structure\}}}}{17}
     {\{report\_structure\_section\}}{{{\color{promptcomment}\itshape\{report\_structure\_section\}}}}{25}
     {\{prior\_report\_structure\}}{{{\color{promptcomment}\itshape\{prior\_report\_structure\}}}}{23}
     {\{reference\_report\_structure\}}{{{\color{promptcomment}\itshape\{reference\_report\_structure\}}}}{26}
     {\{target\_report\_structure\}}{{{\color{promptcomment}\itshape\{target\_report\_structure\}}}}{23}
     {\{target\_report\_structure\_section\}}{{{\color{promptcomment}\itshape\{target\_report\_structure\_section\}}}}{31}
     {\{questions\}}{{{\color{promptcomment}\itshape\{questions\}}}}{11}
     {\{section\_objective\}}{{{\color{promptcomment}\itshape\{section\_objective\}}}}{18},
}

\tcbset{
  promptbox/.style={
    breakable,
    enhanced,
    fonttitle=\scriptsize\bfseries\rmfamily,
    coltitle=black,
    colbacktitle=gray!20,
    colback=gray!3,
    colframe=gray!45,
    boxrule=0.5pt,
    arc=2pt,
    top=3pt, bottom=2pt, left=4pt, right=4pt,
  }
}

This appendix lists the financial data-synthesis prompts used by the pipeline in \S\ref{ssec:data-synthesis}. The academic pipeline follows the same overall structure with domain-specific adjustments for paper HTML, figures, and literature-review style output.

\subsection{Grounding Document Synthesis}
\label{app:prompts-grounding}

These prompts implement the financial grounding-document construction described in \S\ref{ssec:data-synthesis}, turning an iXBRL filing into structured Markdown documents.

\subsubsection{Financial Grounding Documents}
\label{app:prompts-grounding-financial}

\paragraph{File split prompt.}
The below prompt is used to partition the cleaned filing HTML into semantically coherent files (business overviews, risk factors, segment results, financial statements, etc.).


\begin{tcolorbox}[promptbox, title={File Split Prompt (Financial Quarterly)}]
\begin{lstlisting}[style=promptmd]
You are an expert in document analysis and structuring. Your task is to analyze a given SEC filing (HTML format) and reverse-engineer its contents. Your goal is to deconstruct the single HTML document into a set of logically distinct source files, simulating the internal documents a company would prepare before a formal filing.

The final output must be a `FileSet` JSON object containing these reconstructed files, each with its own logical content structure.

# # JSON Output Structure
The output must be a `FileSet` containing a list of `File` objects. The entire output must be a single, valid JSON object conforming to the following Python dataclass definitions:

```python
class SectionElement(BaseModel):
    section_type: Literal["section"]
    heading: str
    objective: str
    children: List["Element"] = Field(default_factory=list)

class TableElement(BaseModel):
    section_type: Literal["table"]
    columns: List[str]
    objective: str

class TextElement(BaseModel):
    section_type: Literal["text"]
    objective: str

Element = Union[SectionElement, TableElement, TextElement]

class File(BaseModel):
    file_name: str
    content: SectionElement

class FileSet(BaseModel):
    files: List[File]
```

# # Instructions
1. **Reconstruct Purpose-Driven Source Files**: Your primary goal is to create a `FileSet` containing several distinct files (ideally 8-12).
    - Each file must represent a logical, standalone source document, not merely a slice of the original SEC filing. Structure each file to serve a specific purpose, as if it were an internal company document. For example, instead of `item-8`, create a file named `annual_financial_statements` to hold financial content.
    - **Aim to balance the length and scope of each file**. If a topic (such as "financial statement notes") contains extensive information, split it into multiple files based on logical subtopics, so that no single file becomes disproportionately large.

2. **Ensure Comprehensive Coverage**: You must process the entire input HTML document.
    - Focus on extracting all substantive information while omitting generic content. Your combined files must include all crucial quantitative data (financial statements, metrics, statistics) and qualitative insights (strategic analysis, risk factors, management's outlook). The content of your proposed files, when combined, must cover all the useful and specific information without any information loss.
    - Exclude non-essential information such as boilerplate legal disclaimers, repetitive headers/footers, and standard corporate governance text that are unlikely to change year-over-year or lacks specific insight.

3. **Structure Content Logically**: Within each file, build a clear hierarchical structure using nested nodes. Ensure every section (`section_type`: "section") is well-balanced.
    - **Do not create sections that consist solely of tables**. Each section must include at least one text element (`section_type`: "text") to introduce, analyze, or provide context for the data before or after any table. This approach ensures each file is coherent and reads fluently.

4. **Define Naming and Objectives**: When creating the `file_name`, `heading`, and `objective`, do not refer back to the SEC filing or its structure. Imagine each file is created for its own internal purpose and has nothing to do with the SEC filing; the filing simply uses these files as sources of raw data. For every element in your structure, write a concise `objective` describing the purpose of that element and what information need to be covered by that element.

5. **Generate a JSON Output**: Construct the final `FileSet` as a single, valid JSON object that adheres to the provided schema.

# # Example JSON Output
Below is an example of the output object structure.

```json
{json_output_example}
```

# # Document for Analysis
```html
{raw_document}
```

Follow the instructions and return me the JSON object. You should only use printable ASCII characters in your response.
\end{lstlisting}
\end{tcolorbox}

\paragraph{Content generation prompt.}
The below prompt is used to generate the full content for each separate grounding document proposed by the file split step. The proposed JSON-format file split for each grounding document is converted into a Markdown file template.


\begin{tcolorbox}[promptbox, title={Content Generation Prompt (Financial Quarterly)}]
\begin{lstlisting}[style=promptmd]
You are an expert document analyst and report writer. Your task is to populate a given Markdown Report Template by accurately extracting information from an HTML Grounding Document.

# # Instructions
1. **Complete the Template**: Your primary task is to find and replace every TODO comment within the Markdown Report Template. Use the provided Grounding Document to find the correct information for each replacement.
    - **Text**: For any TODO comment asking for text (e.g., `TODO: Write text here`), replace the entire comment with clear and factual prose extracted from the source document.
    - **Table**: For any TODO comment before a table (e.g., `TODO: Fill in the table here`), accurately complete the table with the corresponding data from the source document.
    - Ensure the final report is a clean document. All TODO comments must be removed and replaced with the requested information.

2. **TODO Objectives**: Every TODO comment in the Markdown Report Template includes an objective, which describes the purpose and the content that needs to be covered by the TODO. When completing a TODO, you must obtain all relevant information from the grounding HTML document that matches the objective and use it to fulfill the TODO.

3. **Strictly Adhere to Placeholders**: This is a **critical** rule. You must only write content where a TODO comment is present. Do not modify, add to, or delete any of the existing text, headings, or formatting in the template. The template's content, outside of the TODO comments, must remain completely untouched.

4. **Write a Factual, Standalone Report**: All information must come directly from the Grounding Document. However, the final report must be a self-contained document that reads as if it were authored independently. Do not include any phrases that refer back to the source SEC filing, such as "According to the 10-K," "As stated in the filing," or "The document indicates...". The report should stand on its own without any reliance on the source document.

5. **Preserve Structure and Formatting**: You must keep the original structure, headings, and all other Markdown formatting of the template. Your only job is to fill in the blanks designated by TODO.

6. **Final Output**: Your response must be only the completed Markdown report. Do not include any introductory text, explanations, or conversational filler.

# # Document for Analysis
## ## Grounding Document
```html
{raw_document}
```

## ## Markdown Report Template
```markdown
{report_template}
```

Follow the instructions and return me the final report in Markdown output. You should only use printable ASCII characters in your response.
\end{lstlisting}
\end{tcolorbox}

\paragraph{Academic grounding adaptation.}
For arXiv papers, we use an analogous two-stage grounding pipeline: first synthesize a paper outline from HTML, then fill each top-level section separately. The prompt wording is adjusted for academic paper structure, figure references, equations, and literature-review content.

\subsection{Gold Report Synthesis}
\label{app:prompts-gold-report}

These prompts implement the financial gold-report pipeline described in \S\ref{ssec:data-synthesis}. The structure stage has two variants: \emph{form-free} (used at $t_1$) and \emph{form-restricted} (used at $t_2$ and $t_3$, conditioned on the prior report).

\subsubsection{Financial Gold Report Synthesis}
\label{app:prompts-gold-financial}

The financial pipeline merges per-section questions into the structure prompts, so no separate question-augmentation stage is used.

\paragraph{Report structure synthesis (form-free).}
The below prompt is used to propose a hierarchical report structure from the grounding documents alone, applied at $t_1$ when no prior report is available.


\begin{tcolorbox}[promptbox, title={Report Structure Synthesis Prompt -- Form-Free (Financial Quarterly)}]
\begin{lstlisting}[style=promptmd]
You are an expert content strategist specializing in information synthesis. Your task is to analyze a collection of grounding documents, identify the critical information, and propose a cohesive outline for a new, consolidated report. The goal is to synthesize diverse information from the source documents into a single, fluent, and logical narrative.

The grounding documents are provided as a sequence of documents, with each document represented as a Markdown code fence. You must output the proposed report outline as a single JSON object.

# # JSON Output Structure
The output must be a JSON object representing a tree structure. The root must be a `SectionElement` node. Each node in the tree must conform to the following Python dataclass definitions:

```python
class SectionElement(BaseModel):
    section_type: Literal["section"]
    heading: str
    children: List["Element"] = Field(default_factory=list)

class TableElement(BaseModel):
    section_type: Literal["table"]
    columns: List[str]
    objective: str
    questions: List[str] = Field(default_factory=list)

class TextElement(BaseModel):
    section_type: Literal["text"]
    objective: str
    questions: List[str] = Field(default_factory=list)

Element = Union[SectionElement, TableElement, TextElement]
```

# # Instructions
1. **Synthesize and Structure**: Read all the provided grounding documents. Your primary goal is to identify the main themes, key data points, and important narratives. Propose a new report structure that logically combines this information:
    - This structure must be hierarchical and granular, with great readability.
    - Avoid creating long flat sections, instead you should break long sections into nested subsections and continue nesting with sub-subsections, as deep as necessary to keep each part focused.
    - Each top-level section must represent a distinct, high-level theme from your analysis. Inside each section, you could create nested subsections.

2. **Prioritize Importance**: Be selective. You do not need to include every piece of information. Focus only on the significant sections, text, and tables that are essential for the final report. Omit redundant details, boilerplate text, or minor points.
    - A good practice is to identify and structure the report around the **10-12** most important and informative tables from the source material. Focus your outline on these key tables and the corresponding text needed to create a cohesive narrative.

3. **Define Clear Purpose (Crucial)**: For every `TextElement` and `TableElement`, you must define its specific purpose:
    - **objective**: Write a single, clear paragraph describing what this text block or table is meant to achieve. What is its core message? What information should it convey to the reader?
    - **Style**: For the `objective` of `TextElement`, you must also specify the style by beginning the `objective` with one sentence from either "Use descriptive text." or "Combine descriptive text and listing."
    - **questions**: This field is a critical part. Aim to create a comprehensive list of around 5-8 atomic questions that serve as a complete blueprint for generating this node's final content. These questions must cover the entirety of the information required. Your questions could (and not limited to) ask about:
        - Key questions that need to be answered. Frame each question to capture specific metrics or insights required.
        - Required data points to collect.
        - Analysis methods to apply.
        - Metrics to highlight.

4. **Ensure Logical Flow**: The proposed outline should tell a coherent story. Avoid creating sections that only contain tables.
    - Each section should ideally include explanatory `TextElement` nodes to introduce topics and provide context for any `TableElement` nodes that follow.
    - Every `TableElement` must be connected to the narrative through a `TextElement`. A single `TextElement` may introduce, analyze or discuss several tables, but under no circumstances should a table appear without being referenced in an accompanying text block. If you feel a table is not discussed, add a following `TextElement` to discuss it (the objective for this text block could explicitly say "discuss the preceding table...").

5. **Curate Table Columns**: When you include a table in the outline, do not always blindly copy all the columns. For source tables with a large number of columns (e.g., more than 5 or 7), it is better to only select important columns that are referenced or explained in the accompanying text of the grounding documents.

6. **Generate a Single JSON Object**: Construct the final JSON object representing your proposed report structure. Ensure the output is a single, valid JSON object that adheres strictly to the defined Python models.

# # An Example
Below is a simple example including the grounding files and the output JSON object representing the outline of the proposed final report.

```markdown
{markdown_grounding_documents_example}
```

A good output for this example is as follows:

```json
{json_output_example}
```

# # Documents for Analysis
{grounding_documents}

Now, follow all instructions and return the report structure you have designed in a valid JSON format. You should only use printable ASCII characters in your response.
\end{lstlisting}
\end{tcolorbox}

\paragraph{Report structure synthesis (form-restricted).}
The below prompt is used at $t_2$ and $t_3$ to adapt the prior report's structure to the new grounding documents, keeping section layout aligned with the earlier snapshot.


\begin{tcolorbox}[promptbox, title={Report Structure Synthesis Prompt -- Form-Restricted (Financial Quarterly)}]
\begin{lstlisting}[style=promptmd]
You are an expert content strategist specializing in information synthesis. Your task is to analyze a new collection of grounding documents and generate a cohesive report outline in JSON format.

You will be given two inputs:
1. Grounding Documents: A new set of documents to be analyzed.
2. Report Structure Template: A JSON object that serves as the "golden" structure for the output.

Your primary goal is to adapt the Report Structure Template to fit the new Grounding Documents. The final JSON output should try to, **but not forced to**, follow the template's structure, hierarchy, and the intent of its questions and objectives, but with all specific details (like names, dates, metrics, etc.) adjusted and updated to reflect the new information in the new grounding documents.

# # JSON Output Structure
The output must be a single JSON object. Each node in the tree must conform to the following Python dataclass definitions:

```python
class SectionElement(BaseModel):
    section_type: Literal["section"]
    heading: str
    children: List["Element"] = Field(default_factory=list)

class TableElement(BaseModel):
    section_type: Literal["table"]
    columns: List[str]
    objective: str
    questions: List[str] = Field(default_factory=list)

class TextElement(BaseModel):
    section_type: Literal["text"]
    objective: str
    questions: List[str] = Field(default_factory=list)

Element = Union[SectionElement, TableElement, TextElement]
```

# # Instructions
1. **Use the Template as Blueprint**: Start with the provided Report Template as a blueprint, but prioritize creating a logical, informative, and factual report based on the new Grounding Documents.

2. **Adapt Content**: To ensure the report is accurate, factual, and tells a coherent story, you are allowed to make the following changes to the template structure if needed:
    - **Irrelevant Elements**: If a section, table, or text block from the structure template is irrelevant or unsupported by the new documents, replace it with proper content from the new grounding documents.
    - **New Content**: If the new documents contain important major topics or tables that are not in the structure template, you can add new sections, or elements in the report outline to capture that information.
    - For any elements you keep, modify the `heading`, `objective`, `columns`, and `questions` to be specific to the new information. The intent of each element should try to be preserved as much as possible. For instance, if a template question asks, "What was the user engagement rate for the 'Alpha' feature?", and the new documents are about the 'Beta' feature, your adapted question should be "What was the user engagement rate for the 'Beta' feature?".
    - Specifically for the `objective` of `TextElement`, it should specify the style by beginning the `objective` with one sentence from either "Use descriptive text." or "Combine descriptive text and listing."

3. **All Content Must Be Supported by the Documents**: This is the most important rule. Every `heading`, `objective`, table `column`, and specific `question` in your final JSON output must be supported by information present in the new Grounding Documents. Do not invent topics or include details that cannot be verified from the source material. Your report must be a faithful representation of the provided documents.

4. **Generate a Single JSON Object**: Your final output must be a single, valid JSON object, with no explanatory text before or after it.

# # Report Structure Template
```json
{report_structure_template}
```

# # Documents for Analysis
{grounding_documents}

Follow the instructions and return me the report structure you design in JSON output. You should only use printable ASCII characters in your response.
\end{lstlisting}
\end{tcolorbox}

\paragraph{Final content generation prompt.}
The below prompt is used to fill each section placeholder by conditioning the LLM on the full grounding set and the synthesized structure.


\begin{tcolorbox}[promptbox, title={Final Content Generation Prompt (Financial Quarterly)}]
\begin{lstlisting}[style=promptmd]
You are an expert content strategist specializing in information synthesis. Your task is to populate a given Markdown Report Template by accurately extracting information from a collection of grounding documents.

The grounding documents are provided as a sequence of documents, with each document represented as a Markdown code fence.

# # Instructions
1. **Complete the Template**: Your primary task is to find and replace every TODO comment within the Markdown Report Template. Use the provided Grounding Document to find the correct information for each replacement.
    - **Text**: For any TODO comment asking for text (e.g., `TODO: Write text here`), replace the entire comment with clear and factual prose extracted from the source document.
    - **Table**: For any TODO comment before a table (e.g., `TODO: Fill in the table here`), accurately complete the table with the corresponding data from the source document.
    - Ensure the final report is a clean document. All TODO comments must be removed and replaced with the requested information.

2. **TODO Objectives**: Every TODO comment in the Markdown Report Template includes an objective and a list of questions.
    - **objective**: This describes the purpose and the content that needs to be covered by the TODO.
        - For text, if the objective says "Use descriptive text.", then you should write fluent paragraphs of text content, and do not use lists in the Markdown. If the objective says "Combine descriptive text and listing.", then you can use lists to structure the text content.
    - **questions**: This is your checklist. Your generated text or table data should cover the answer of every single question in this list. Use it to guide your data extraction, analysis, and writing. However, if you cannot find or derive the answer of a question from the grounding documents, you can skip that question.
    - You do not need to write a paragraph for each question one by one. Instead, skillfully combine the answers of multiple questions into natural, cohesive paragraphs. Aim to combine short paragraphs on similar topics into long paragraphs to increase the readability.
    - Once you complete the content, do not include the TODO comments in the final output.

3. **Strictly Adhere to Placeholders**: This is a **critical** rule. You must only write content where a TODO comment is present. Do not modify, add to, or delete any of the existing text, headings, or formatting in the template. The template's content, outside of the TODO comments, must remain completely untouched.

4. **Write a Factual, Standalone Report**: All information must come directly from the Grounding Documents. However, the final report must be a self-contained document that reads as if it were authored independently.
    - Do not include any phrases that refer back to the source grounding documents, such as "As stated in the document," or "The document indicates...". The report should stand on its own without any reliance on the source document.
    - Do not blindly copy paragraphs from the grounding documents; rephrase the text properly to make the final report natural in its own flow.
    - **All Content Must Be Supported by the Documents**: This is the most **important** rule. Every content in your final output must be supported by information present in the Grounding Documents. Do not invent topics or include details that cannot be verified from the source material. Your report must be a faithful representation of the provided documents.

5. **Final Output**: Your response must be only the completed Markdown report. Do not include any introductory text, explanations, or conversational filler.

# # Document for Analysis
## ## Grounding Documents
{grounding_documents}

## ## Markdown Report Template
```markdown
{report_template}
```

Follow the instructions and return me the final report in Markdown output.
\end{lstlisting}
\end{tcolorbox}

\paragraph{Academic gold-report adaptation.}
For academic reports, we follow the same form-free and form-restricted synthesis pattern, with prompt adjustments for survey-style organization, paper comparison, multimodal evidence, and a separate question-augmentation step for longer sections.

\end{document}